\documentclass[10pt]{article}

\usepackage[preprint]{tmlr}

\usepackage[utf8]{inputenc}
\usepackage[T1]{fontenc}
\usepackage{url}
\usepackage{booktabs}
\usepackage{amsfonts}
\usepackage{amsmath, amssymb, amsthm, mathtools}
\usepackage{algorithm}
\usepackage{algpseudocode}
\usepackage{enumitem}
\usepackage{microtype}
\usepackage{graphicx}
\usepackage{hyperref}
\hypersetup{hidelinks}
\usepackage{xcolor}
\usepackage{needspace}

\newtheorem{theorem}{Theorem}
\newtheorem{proposition}[theorem]{Proposition}
\newtheorem{lemma}[theorem]{Lemma}
\newtheorem{corollary}[theorem]{Corollary}
\theoremstyle{definition}
\newtheorem{assumption}{Assumption}
\newtheorem{remark}{Remark}
\newtheorem{definition}{Definition}

\newcommand{\R}{\mathbb{R}}

\newcommand{\F}{\mathcal{F}}
\newcommand{\NRMSE}{\mathrm{NRMSE}}
\newcommand{\FTLE}{\mathrm{FTLE}}
\newcommand{\CV}{\mathrm{CV}}

\newcommand{\CVeff}{\CV_{\mathrm{eff}}}

\title{Rollout Scheduling for Chaotic Neural ODEs:
Coverage Gains, Computational Cost, and Predictive Accuracy}

\author{\name Ajith Nayanananda \email anayananad@odu.edu \\
      \addr Department of Mathematics and Statistics\\
      Old Dominion University, Norfolk, VA 23529
      \AND
      \name Xuping Xie\thanks{Corresponding author.} \email xxie@odu.edu \\
      \addr Department of Mathematics and Statistics\\
      Old Dominion University, Norfolk, VA 23529}

\def\month{MM}  
\def\year{YYYY} 
\def\openreview{\url{https://openreview.net/forum?id=XXXX}} 

\begin{document}
\maketitle

\begin{abstract}
Training Neural Ordinary Differential Equations (Neural ODEs)
on chaotic systems is difficult because small changes in model parameters
can lead to rapidly growing differences in predictions. A rollout is a
trajectory predicted by the model. Choosing its length from a single
global Lyapunov time overlooks differences in local instability.
We study rollout scheduling for training on one
contiguous rollout per sampled state, asking whether adapting
its length to local instability improves prediction. We find the schedule that uses the most trajectory data under an instability budget, a limit on how much instability any one rollout may accumulate. Under stated assumptions, we bound the resulting coverage gain and the coverage lost when the instability estimates are in error.
Numerical checks illustrate the theory. Training experiments on
Lorenz-63, Lorenz-96, and Kuramoto--Sivashinsky compare adaptive and
constant schedules at the same instability budgets and maximum
rollout lengths. The budgets are computed from the true system's
finite-time Lyapunov exponents. Across the tested settings, adaptive
schedules increase trajectory coverage at additional computational
cost, without showing a consistent improvement in prediction accuracy.
On Lorenz-63, limiting adaptive rollouts to the horizon used to estimate
instability improves prediction at the larger budgets. On
Kuramoto--Sivashinsky, most configurations provide useful short-time
predictions, while all tested long rollouts eventually fail.
These findings show why coverage, prediction
accuracy, and long-time behavior need separate evaluation, and why
instability estimates should be checked over the windows used in training.
\end{abstract}

\section{Introduction}

Learning chaotic dynamics requires choosing how far into the future a model predicts
during each training step. Neural Ordinary Differential Equations
(Neural ODEs) \citep{chen2018node} learn a vector field by integrating
trajectories and comparing them with observations. We call each
predicted trajectory a rollout and its duration the window length.
Longer windows expose the model to more trajectory data, but require
more computation and can make predictions more sensitive to changes
in model parameters
\citep{mikhaeil2022difficulty,jiang2023invariant,park2024statistically}. This suggests using shorter windows where nearby trajectories
separate quickly and longer windows where they separate more slowly.
Finite-time Lyapunov exponents (FTLEs) measure how fast small differences
in starting states can grow over a chosen time interval. These rates
vary across the states of a chaotic system
\citep{prasad1999ftle,shibata1999ks,ott2002chaos}; under a common
instability budget, that is, a shared limit on the integrated
local growth rate that any single rollout may accumulate, this
variation can allow a greater total trajectory length at each training
update. We call that length
\emph{coverage}. It includes overlapping segments and does not measure
independent information or model quality. Throughout, the
budget constrains instability, not computation: schedules at the same
budget can differ several-fold in solver work, and we report
computational cost separately.

We ask whether assigning a separate rollout length to each sampled
state improves prediction enough to justify its additional cost,
compared with constant windows that share the same
instability budget, computed from the true system's finite-time
Lyapunov exponents, and the same horizon cap. We also test how performance changes when
the cap extends beyond the FTLE measurement horizon. Our analysis finds the greatest coverage possible for
schedules that train on one contiguous rollout per sampled
state rather than on several penalty-coupled pieces.
Letting each segment's window grow until its integrated
instability reaches the budget
or the window reaches its maximum allowed length gives the
greatest coverage within these limits (Theorem~\ref{thm:optimality}).
When growth rates are constant within each window, a lower
bound relates variation in the instability profiles
to the coverage gain, provided no window is changed by the
minimum or maximum length (Corollary~\ref{cor:gap}). Additional
bounds describe how instability affects gradients and how estimation
errors affect coverage, under stated assumptions
(Lemma~\ref{lem:sensitivity}, Proposition~\ref{prop:robust}). Numerical
checks on four systems illustrate these results without model training.

A separate study of $305$ runs on three systems, with five seeds per
configuration, tests whether greater coverage improves
prediction at
shared budgets and caps. Adaptation uses more trajectory data at
additional cost without showing a consistent improvement in prediction;
the paired confidence intervals also do not show that
the schedules give equal prediction accuracy. We also compare against fixed windows and a penalty-coupled baseline, which do show accuracy differences between schedules. The cap and long-time evaluations show further limits. A smaller
cap removes the drop in prediction accuracy observed on Lorenz-63 at
the larger budgets when adaptive windows extend beyond the FTLE
measurement horizon. Checks of the instability estimates suggest a
possible explanation, but the cap also changes window lengths, loss
weights, computation, and sampled starting states. Every tested long
Kuramoto--Sivashinsky (KS) rollout fails
despite useful short-time prediction for most configurations.
These comparisons separate the amount of trajectory used in
training from the quality of the resulting model.

\section{Background and related work}
\label{sec:2-background}

A Neural ODE models $\dot x=f_\theta(x)$ with learned flow
$x_\theta(t;x_0)$. Its rollout loss is
\begin{equation}
\mathcal{L}_{\mathrm{roll}}(\theta;W)=\sum_{j=0}^m\omega_j\,
\ell(x_\theta(t_j;x_0),x^\star(t_j;x_0)),
\label{eq:roll}
\end{equation}
where $x_0$ is the initial state, $\ell$ is a pointwise loss
such as the squared error between the two states, and the
nonnegative quadrature weights $\omega_j$ and nodes
$0=t_0<\cdots<t_m\le W$ cover $[0,W]$ and
$x^\star$ is the reference trajectory. Training may use a single
rollout per sampled state, as studied here, or
several shorter windows coupled by discontinuity
penalties, which split the window into penalized pieces.
The penalties add to the loss when the end of one predicted
piece does not match the start of the next.

The horizon also determines the interval over which instability
affects the rollout. For a chaotic flow $\varphi^t$, the largest Lyapunov exponent
$\lambda_1>0$ defines the timescale $T_\lambda=1/\lambda_1$.
Finite-window stretching is measured by
\begin{equation}
\lambda_{\FTLE}(x_0,\tau)=\tfrac1\tau
\log\sigma_{\max}(D\varphi^\tau(x_0)).
\label{eq:ftle}
\end{equation}
Here $D\varphi^\tau(x_0)$ is the Jacobian of the time-$\tau$ flow
map with respect to the initial state, the tangent matrix, and
$\sigma_{\max}$ denotes its largest singular value, so the logarithm
measures the maximal stretching of an infinitesimal perturbation over
the horizon $\tau$.
Its variation across the attractor motivates local scheduling
\citep{prasad1999ftle,shibata1999ks}. Parameter sensitivities follow
a variational equation; Lemma~\ref{lem:sensitivity} bounds their effect
on the gradient through integrated growth and additional prefactors.
The FTLE of the true system that generated the data supplies
a measurable proxy for the learned model's instability,
whose relation to that learned-model bound requires separate assessment.

Lyapunov-guided training horizons have established precedents.
\citet{mikhaeil2022difficulty} relate chaotic dynamics to gradient
growth and choose sparse teacher-forcing intervals using the largest
Lyapunov exponent. MP-NODE \citep{chakraborty2024mpnode} couples short windows with
discontinuity penalties, using a global Lyapunov estimate to choose
their lengths; WP-NODE \citep{li2025weak} uses weak-form residuals on
fixed partitions. Multiple shooting and homotopy methods also couple
windows \citep{turan2022multiple,iakovlev2023shooting,ko2023homotopy},
while recent approaches progressively merge them
\citep{persson2026cms} or extend discontinuity penalties to neural PDE
surrogates \citep{chakraborty2024mplosses}. These methods extend
beyond the scope of our theory for the one-rollout-per-state setting
studied here. Other approaches change the data or dynamics used in training.
CCF \citep{vejendla2025ccf} selects trajectories using global dynamical
quantities, and generalized teacher forcing \citep{hess2023gtf} mixes
teacher and model samples at a fixed horizon. LyaNet
\citep{rodriguez2022lyanet}, stabilized NODE \citep{linot2023stabilized},
and gradient flossing \citep{engelken2023flossing} regularize dynamics;
invariant-measure objectives target long-time statistics
\citep{jiang2023invariant,park2024statistically,schiff2024dyslim}.
Related work uses FTLEs for post-hoc network analysis \citep{storm2024ftle} or
classifier robustness \citep{kuehn2026tracking}, and evaluates
forecasting curricula \citep{teutsch2022flipped}.

Rollout length also matters in neural PDE and weather prediction.
GraphCast increases training unrolls from one to twelve steps
\citep{lam2023graphcast}; \citet{list2025unrolled} examine unrolling,
gradient propagation, and distribution shift. The pushforward trick
trains on model-generated inputs while stopping gradients through the
preceding prediction \citep{brandstetter2022mp}, whereas PDE-Refiner
denoises within each temporal prediction \citep{lippe2023pderefiner}.
Related choices include backpropagation truncation in recurrent
chaotic forecasters \citep{vlachas2020backprop} and prescribed forecast
targets in horizon forcing \citep{zhuang2025horizon}. The hybridizable
integrator of \citet{kinch2026hybridizable} obtains conditional gradient
bounds through a structure-preserving discretization.

Prior work establishes that rollout length affects training
\citep{mikhaeil2022difficulty,list2025unrolled} and that short-time
accuracy does not ensure correct long-time behavior
\citep{park2024statistically,schiff2024dyslim}. Our specific contribution
is to formalize and evaluate state-dependent allocation under a common
instability budget. Budgets computed from the true system's
FTLE, matched between schedules together with the horizon cap,
set the conditions for the training comparison. We measure
the resulting coverage, computational cost, and prediction accuracy.
The cap comparison and measurements of trajectory growth also bear on
whether instability measured over a short window describes what
happens over the longer windows used in training.

\section{Instability-constrained rollout scheduling}
\label{sec:theory}

\subsection{Setup and sensitivity bound}

Assume full-state training trajectories $x^\star_n(t)$ are generated by an
unknown chaotic system, and let $s_i$ denote a sampled start time on one
such trajectory. We train a Neural ODE on segments of variable duration
$W_i \in [W_{\min}, W_{\max}]$, and we call the upper limit
$W_{\max}$ the horizon cap. For segment $i$, define
\begin{equation}
\mathcal{L}^{(i)}_{\mathrm{roll}}(\theta; W_i) =
\sum_{j=0}^{m_i} \omega_{i,j}\,
\ell\bigl(x_\theta(t_{i,j}; x^\star_{n}(s_i)),\, x^\star_n(s_i + t_{i,j})\bigr),
\label{eq:segloss}
\end{equation}
with $m_i+1$ quadrature nodes $0 = t_{i,0} < \cdots < t_{i,m_i} \le W_i$
and weights $\omega_{i,j}$.
Let $S_i(t) = \partial_\theta x_\theta(t; x^\star_n(s_i))$ denote the
parameter sensitivity matrix; it satisfies
\begin{equation}
\dot S_i(t) = \partial_x f_\theta(x_\theta(t))\, S_i(t)
+ \partial_\theta f_\theta(x_\theta(t)),
\qquad S_i(0) = 0.
\end{equation}
Write $\Phi_i(t,s)$ for the state-transition matrix of the homogeneous
variational equation $\dot z(t) = \partial_x f_\theta(x_\theta(t))\, z(t)$,
so that
\begin{equation}
S_i(t) = \int_0^t \Phi_i(t, s)\, \partial_\theta f_\theta(x_\theta(s))\, ds.
\label{eq:varofconst}
\end{equation}

\label{sec:sensitivity}

The following bound explains why we limit the instability
accumulated during a rollout: it relates that instability to the
gradient used in training. The bound alone does not determine whether
adaptive windows improve prediction.

\begin{assumption}[Local growth and Lipschitz constants]
\label{assump:bounds}
For segment $i$ the quadrature weights satisfy $\omega_{i,j} \ge 0$, and
there exist a measurable local growth-rate proxy
$\mu_i: [0, W_i] \to \R_{\ge 0}$ and constants $C_\Phi, B_\theta, B_\ell > 0$
such that, for all $0 \le s \le t \le W_i$ and $j = 0, \ldots, m_i$,
with $\|\cdot\|$ the operator norm induced by the Euclidean norm,
\begin{align}
\|\Phi_i(t, s)\| &\le C_\Phi \exp\!\left(\int_s^t \mu_i(r)\,dr\right),
\label{eq:phibound}\\
\|\partial_\theta f_\theta(x_\theta(t))\| &\le B_\theta,\\
\|\partial_x \ell(x_\theta(t_{i,j}), x^\star(t_{i,j}))\| &\le B_\ell.
\end{align}
\end{assumption}

\begin{lemma}[Sensitivity bound]
\label{lem:sensitivity}
Under Assumption~\ref{assump:bounds},
\begin{equation}
\bigl\|\nabla_\theta \mathcal{L}^{(i)}_{\mathrm{roll}}\bigr\|
\;\le\; C_\Phi B_\theta B_\ell \sum_{j=0}^{m_i} \omega_{i,j}
\int_0^{t_{i,j}} \exp\!\left(\int_s^{t_{i,j}} \mu_i(r)\,dr\right)\,ds.
\label{eq:gradbound}
\end{equation}
In particular, if $\Lambda_i(W_i) := \int_0^{W_i} \mu_i(r)\,dr \le \beta$,
then
\begin{equation}
\bigl\|\nabla_\theta \mathcal{L}^{(i)}_{\mathrm{roll}}\bigr\|
\;\le\; C_\Phi B_\theta B_\ell \,\Bigl(\sum_{j=0}^{m_i} \omega_{i,j}\, t_{i,j}\Bigr) e^\beta.
\label{eq:gradbudget}
\end{equation}
\end{lemma}

The proof is in Appendix~\ref{app:proof}.

\begin{remark}[Recovery of the global heuristic]
\label{rem:globalspecial}
If $\mu_i(t) \equiv \bar\lambda$ is constant, then $\Lambda_i(W_i) = \bar\lambda W_i$,
so the budget rule $\Lambda_i(W_i) \le \beta$ reduces to the familiar global
constraint $W_i \le \beta / \bar\lambda$. The original ``window proportional
to Lyapunov time'' heuristic is the constant-$\mu$ special case of the
present formulation.
\end{remark}

\subsection{Coverage-optimal allocation}
\label{sec:optimality}

Consider a batch of $B$ sampled segments with cumulative instability
profiles $\Lambda_i(W) = \int_0^W \mu_i(r)\,dr$, $i = 1, \ldots, B$. Each
$\Lambda_i$ is non-decreasing and continuous in $W$. Each
profile records the local growth rate integrated over a window of
length $W$, measured in nats (natural-log units); a budget $\beta$ is
an upper limit on this integrated quantity. A
\emph{rollout schedule} is a vector $W = (W_1, \ldots, W_B) \in
[W_{\min}, W_{\max}]^B$.

\begin{definition}[Coverage and integrated instability]
\label{def:coverage}
For a schedule $W$,
\[
C(W) := \sum_{i=1}^B W_i, \qquad
G(W) := \max_{1 \le i \le B} \Lambda_i(W_i).
\]
$C(W)$ is the total trajectory length the learner integrates over;
$G(W)$ is the worst-case integrated instability across the batch and, by
Lemma~\ref{lem:sensitivity}, enters the gradient bound through
$e^{G(W)}$, alongside the stated prefactors and quadrature terms.
\end{definition}

\begin{definition}[Budget-feasible set]
For $\beta \ge 0$, the budget-feasible set is
\[
\F(\beta) := \{ W \in [W_{\min}, W_{\max}]^B : G(W) \le \beta \}.
\]
\end{definition}

\begin{definition}[Saturating horizon]
For each segment $i$ and budget $\beta$, the segment-wise saturating
horizon, the longest window whose integrated instability stays
within the budget, is
\begin{equation}
W_i^\star(\beta) := \max\!\left\{ W \in [W_{\min}, W_{\max}] : \Lambda_i(W) \le \beta \right\},
\label{eq:satwidth}
\end{equation}
which exists by continuity of $\Lambda_i$. Set $W^\star(\beta) :=
(W_1^\star(\beta), \ldots, W_B^\star(\beta))$.
\end{definition}

We assume $\beta$ is large enough that $\Lambda_i(W_{\min}) \le \beta$ for
all $i$, so the feasible set is nonempty; otherwise $W_{\min}$ must be
reduced or the budget increased to make it nonempty.

\begin{theorem}[Coverage optimality under an instability budget]
\label{thm:optimality}
Fix a budget $\beta \ge 0$. Then:
\begin{enumerate}
    \item[(i)] $W^\star(\beta) \in \F(\beta)$.
    \item[(ii)] $W^\star(\beta)$ is the unique maximizer of $C$ on $\F(\beta)$.
    \item[(iii)] For any global schedule $W^g = (W_g, \ldots, W_g)$ with
    $W^g \in \F(\beta)$,
    \[
    W_g \le W^\star_{\min}(\beta) := \min_i W_i^\star(\beta),
    \qquad
    C(W^\star) - C(W^g) = \sum_{i=1}^B \bigl( W_i^\star(\beta) - W_g \bigr) \ge 0,
    \]
    with equality if and only if $W_i^\star(\beta) = W_g$ for every $i$. In
    particular, against the largest feasible global window
    $W_g = W^\star_{\min}(\beta)$ the gap is zero if and only if all
    saturating horizons coincide, and positive otherwise.
\end{enumerate}
\end{theorem}

Every allowed window satisfies $W_i\le W_i^\star$, so choosing
these longest windows gives the greatest total coverage. Full proofs
are in Appendix~\ref{app:proof}. The adaptive rule covers more trajectory data
than the largest allowed constant window exactly when the
longest allowed windows differ. Different instability
profiles can still give equal window lengths: they may reach the
budget at the same time, or all windows may reach the maximum allowed
length first.

\begin{corollary}[Heterogeneity gap]
\label{cor:gap}
Let $W_g^{\max} := \min_i W_i^\star(\beta)$ be the largest feasible global
window, and let $\mathbf{1}$ denote the all-ones vector in $\R^B$,
so that $W_g^{\max}\mathbf{1}$ is the constant schedule. Then the relative coverage gain of the adaptive rule over $W_g^{\max}$
is
\begin{equation}
R(\beta) := \frac{C(W^\star) - C(W_g^{\max} \mathbf{1})}{C(W_g^{\max}\mathbf{1})}
= \frac{\bar W^\star - W_g^{\max}}{W_g^{\max}},
\qquad \bar W^\star := \frac{1}{B}\sum_i W_i^\star(\beta).
\label{eq:gapratio}
\end{equation}
This ratio is non-negative and vanishes if and only if all saturating
horizons agree. Suppose in addition that the profiles are linear on the
relevant range, $\Lambda_i(W) = \mu_i W$ with effective rates $\mu_i > 0$,
and that no horizon is clipped by $[W_{\min}, W_{\max}]$, so that
$W_i^\star(\beta) = \beta/\mu_i$. Then
\begin{equation}
R(\beta) \;\ge\; \CVeff^2, \qquad
\CVeff := \frac{\operatorname{std}_i(\mu_i)}{\operatorname{mean}_i(\mu_i)},
\label{eq:cvbound}
\end{equation}
where the statistics are taken over the batch. The bound is deterministic
and holds for every finite batch.
\end{corollary}

The proof is in Appendix~\ref{app:cor-proof}. The bound is
one-sided: a large $\CVeff$ guarantees a large gap under these
assumptions, but a small $\CVeff$ does not bound the gap from above.
The identity $R+1=\max_i\mu_i\,\overline{1/\mu}$ shows that the gap
depends on extreme rates. Here $\overline{1/\mu}=\tfrac1B\sum_i 1/\mu_i$
is the batch mean of the reciprocal rates.
For the schedule using FTLEs, $\mu_i=\max\{\widehat\lambda_{\FTLE,i},L\}$
replaces the estimated FTLE $\widehat\lambda_{\FTLE,i}$ of
segment $i$ by a positive floor $L$ whenever the estimate is smaller;
this floor is equivalently a cap $W_{\max}=\beta/L$.
The inequality concerns this effective rate's coefficient of
variation (CV, the ratio of standard deviation to mean), not the raw FTLE CV;
both are reported in the numerical checks.

\subsection{Robustness to estimation error}
\label{sec:robustness}

Theorem~\ref{thm:optimality} assumes exact profiles. With estimated
profiles, Proposition~\ref{prop:robust} gives sufficient conditions
for feasibility under the true instability budget and bounds the
coverage loss relative to the oracle schedule, the schedule
computed from exact profiles.

\begin{proposition}[Robustness of the adaptive rule to estimation error]
\label{prop:robust}
Let $\widehat{\Lambda}_i: [W_{\min}, W_{\max}] \to \R_{\ge 0}$ be a
non-decreasing continuous estimator of $\Lambda_i$, and suppose
\begin{equation}
\bigl|\widehat{\Lambda}_i(W) - \Lambda_i(W)\bigr| \le \epsilon
\qquad \forall\, W \in [W_{\min}, W_{\max}],\ i = 1, \ldots, B,
\label{eq:eps-bound}
\end{equation}
for some uniform error bound $\epsilon \ge 0$. Define the
\emph{$\epsilon$-shrunken} estimated schedule
\begin{equation}
\widetilde{W}_i(\beta; \epsilon) := \max\!\left\{ W \in [W_{\min}, W_{\max}] :
\widehat{\Lambda}_i(W) \le \beta - \epsilon \right\},
\label{eq:tildeW}
\end{equation}
and assume $\beta \ge \epsilon + \widehat{\Lambda}_i(W_{\min})$ for all $i$
so that $\widetilde W_i$ is well-defined. Then:
\begin{enumerate}
    \item[(i)] \emph{(Feasibility under the true budget.)}\,
    $\widetilde W(\beta;\epsilon) \in \F(\beta)$; equivalently,
    $\Lambda_i(\widetilde W_i) \le \beta$ for every $i$.
    \item[(ii)] \emph{(Coverage gap.)}\, Suppose in addition that each
    profile is $L_i$-strongly increasing on $[W_{\min}, W_{\max}]$, in the
    sense that $\mu_i(t) \ge L_i > 0$ for almost every $t$. Then for every
    $i$, either $\widetilde W_i = W_{\max}$ (in which case
    $W_i^\star(\beta) = \widetilde W_i$ and the gap is zero) or
    \begin{equation}
    0 \le W_i^\star(\beta) - \widetilde W_i(\beta;\epsilon) \le \frac{2\epsilon}{L_i},
    \label{eq:per-seg-gap}
    \end{equation}
    and consequently
    \begin{equation}
    C\bigl(W^\star(\beta)\bigr) - C\bigl(\widetilde W(\beta;\epsilon)\bigr)
    \;\le\; 2\epsilon \sum_{i=1}^B \frac{1}{L_i}.
    \label{eq:total-gap}
    \end{equation}
\end{enumerate}
\end{proposition}

The proof is in Appendix~\ref{app:robust-proof}. The bound on lost coverage
scales linearly with profile error and inversely with the lower growth
rates $L_i$. Applying it requires a uniform error bound, budget
shrinkage, and the stated feasibility conditions; true-system estimates or
a gradually increasing budget (the warm-up of
Section~\ref{sec:procedure}) alone do not supply these conditions.
Section~\ref{sec:pilot:ftle} checks the proposition
numerically on all four benchmark systems (Table~\ref{tab:prop4}).

\subsection{The schedule in practice}
\label{sec:procedure}
\label{sec:bridge}
\label{sec:notsay}
\label{sec:caveats}

The training study implements the one-rollout-per-state rule
using estimated instability, window lengths rounded to whole
model steps, and groups of windows with the same length. Throughout,
the budget is computed from the FTLE of the true system that generated
the data, which stands in for the learned model's unknown instability.
We call this value the \emph{nominal} budget: it is the value the proxy
assigns to a window, not a certified bound on the learned model's
sensitivity. The comparisons use the same
nominal budgets and caps, rather than exact learned-model
sensitivities or accuracy at fixed compute. Schedules that
split the window into penalized pieces remain outside the theorem's feasible set. For a sampled start $x^\star_n(s_i)$ and estimated profile
$\hat\mu_i(t)$, the ideal training horizon is
\begin{equation}
W_i = \max\!\left\{ W \in [W_{\min}, W_{\max}] : \int_0^W \hat\mu_i(r)\,dr \le \beta_k \right\},
\label{eq:Wi-rule}
\end{equation}
where a budget warm-up raises the budget gradually with optimizer update $k$ according to
\begin{equation}
\beta_k = \beta_{\min} + (\beta_{\max} - \beta_{\min})\bigl(1 - e^{-k/k_0}\bigr).
\label{eq:betaschedule}
\end{equation}
Here $\beta_k$ is the limit we choose at update $k$, while
$\hat\mu_iW_i$ is the estimated amount of instability used by a
particular window.
In the experiments, $\beta_{\min}=0.25\beta_{\max}$ and $k_0$ is
$10\%$ of the update ceiling, the maximum number of training updates. The implemented step count is
$N_i=\max\{5,\lfloor W_i/\Delta t_{\rm m}\rfloor\}$, where
$\Delta t_{\rm m}$ is the model step. Rounding down preserves
feasibility when the ideal window is feasible; the five-step minimum
can violate the budget, and those cases are audited.
\label{sec:estimators}
All training runs use the FTLE of the true system,
\begin{equation}
\widehat{\lambda}_{\FTLE,i}(\tau) = \frac{1}{\tau} \log \sigma_{\max}\bigl(D\varphi^\tau(x^\star_n(s_i))\bigr),
\label{eq:ftle-estimator}
\end{equation}
computed by propagating the full tangent matrix with periodic scalar
rescaling and taking its largest singular value at $\tau$.
This measures maximal endpoint stretch, unlike the stretch of one
random tangent direction (Appendix~\ref{app:diagnostics}). We use
$\hat\mu_i(t)=\max\{\widehat\lambda_{\FTLE,i}(\tau),L\}$,
where training sets $L$ to the larger of $10^{-3}$ and the smallest
positive training FTLE. The separate coverage checks use percentile
floors. Appendix~\ref{sec:pilot:lemma1} reports running endpoint
stretches, and Appendix~\ref{app:bootstrap} describes an unevaluated
model-based alternative.

We fix the measurement horizon at $\tau=T_\lambda/2$ so that
the instability estimate can be computed before the window length is
chosen. An estimate based on the start and end of this interval can
miss larger growth within it, and its accuracy over longer windows is
not guaranteed. It also describes the true system
rather than the learned model. We therefore test different
values of $W_{\max}/\tau$. Neither keeping windows within $\tau$ nor
observing a large variation in rates across starting states (a large
CV) establishes the error bound required by Proposition~\ref{prop:robust}.
\label{sec:bootstrap}
To limit the number of distinct window lengths in a batch, we
sort the window lengths into $q=4$ groups and use the shortest window in each group.
This cannot increase the estimated instability of any segment. Using
the mean length of each group instead gave a
worst-case batch budget $\Lambda_{\max}:=\max_i\hat\mu_iW_i=2.6$ at nominal $\beta=1$ in a preliminary check.
Each segment's mean-squared rollout error is weighted by $\bar W/W_i$, where $\bar W$ is the mean window length
of the batch, reducing the relative weight of longer windows.
An unweighted ablation assesses this choice, which does not cancel
the sensitivity factors in Lemma~\ref{lem:sensitivity}.
Algorithm~\ref{alg:adaptive} summarizes the smaller-cap implementation.

\section{Experiments}
\label{sec:pilot}
Numerical checks first examine the coverage gain and the
robustness to estimation error, under the stated assumptions, on four systems. Training experiments then
measure coverage, cost, and prediction on three systems, with cap
comparisons and long-time evaluation testing the scope of the findings.

\subsection{Setup}
\label{sec:setup}
\label{sec:experiments}
We use Lorenz-63 (L63) \citep{lorenz1963}, R\"ossler \citep{rossler1976},
Lorenz-96 (L96) at $N{=}40$ \citep{lorenz1996}, and Kuramoto--Sivashinsky
(KS) at $L{=}22$, $N{=}64$. The four systems span
an FTLE $\CV{\in}[0.11, 0.75]$ at
$\tau = T_\lambda/2$. The working reference values
$\lambda_1^{\rm ref}$ in Table~\ref{tab:hetero} define
$T_\lambda=1/\lambda_1^{\rm ref}$ and all normalized horizons.
R\"ossler appears only in the numerical checks of the theory;
the three training systems also have a data estimate of the asymptotic
$\lambda_1$ for comparison with the learned models.

The coverage checks compare the adaptive schedule with the largest
feasible constant window on $4000$ attractor samples per system at
three FTLE horizons, including a sweep of the instability floor.
Separate checks inject bounded errors to test
Proposition~\ref{prop:robust}, use Gaussian errors as a stress test,
and measure true-system endpoint stretch. These calculations involve
no learned model (Appendix~\ref{app:diagnostics}). On Lorenz-63, Lorenz-96 and KS we fix the
model and training protocol within each system and compare five
schedule families, with five seeds per configuration. Constant windows of $W \in \{0.1, 0.25, 0.5, 1,
2\}\,T_\lambda$ (up to $1\,T_\lambda$ on KS) span the instability-budget axis. The
constant window at the same budget uses $W_g^{\max}(\beta) = \beta /
\max_{i \in \text{batch}} \hat\mu_i$, for $\beta \in \{0.5, 1, 2\}$,
before clipping to the window bounds. When the minimum window is
feasible, this gives the largest feasible constant window in
Theorem~\ref{thm:optimality}(iii), subject to discretization.
Here ``constant'' means that every segment in a batch uses the
same window length. That length can change between training updates as
the budget and the sampled states change. This differs from the constant
windows of $0.1$ to $2T_\lambda$ listed above, whose lengths stay the
same throughout training.
The
adaptive rule $W_i^\star(\beta)$ is run at the same three budgets with
the same warm-up $\beta_k$ as its constant counterpart. Both rules are run
with the maximum horizon $W_{\max}$ set to the FTLE horizon $\tau =
T_\lambda/2$, so that the profile is never extrapolated, and on
Lorenz-63 and Lorenz-96 also with $W_{\max} = 2T_\lambda$, which
extrapolates it four times beyond $\tau$; every pair compared in this
paper shares its budget, its warm-up and its $W_{\max}$.

An ablation of the adaptive rule at $W_{\max} = \tau$ drops the
inverse-window loss weighting of Section~\ref{sec:bootstrap}
(``unweighted''). The
global-Lyapunov ramp raises a constant window from $0.1$ to
$1\,T_\lambda$ over training. MP-NODE \citep{chakraborty2024mpnode} is
run with $K \in \{2, 3, 5\}$ chunks of a $1\,T_\lambda$ window ($K \in
\{3, 5\}$ on KS), each chunk re-initialized from data with a unit-weight
end-of-chunk discontinuity penalty. This is $24$ schedules and $120$
runs on each of Lorenz-63 and Lorenz-96 and $13$ schedules and $65$ runs
on KS, where windows beyond $1\,T_\lambda$ and the extrapolated rules
were not run because of computational limits. Every run logs the realized worst-case budget
$\Lambda_{\max} = \max_i \hat\mu_i W_i$ of the batch and the
pre-clipping gradient norm every ten updates and at the final update.
We audit the fraction of logged batches whose
worst segment exceeds the nominal budget, including violations caused
by the five-step minimum (Section~\ref{sec:study:matched}).

For each system one long post-transient orbit is split into
disjoint training, validation, reference, and test windows, and the
true-system FTLE at $\tau=T_\lambda/2$ is precomputed along the
training window. The model is fixed across schedules within each
system: a multilayer perceptron on Lorenz-63 and a periodic
one-dimensional convolutional field on Lorenz-96 and KS, integrated
with fixed-step RK4 at the data step (on KS at twice the data step,
with the loss on every second data point, which halves the number
of integration steps for a rollout of the same duration). Training uses AdamW with the same settings for every schedule
within a system and a minimum window $W_{\min}$ of five model steps.
The checkpoint with the best validation rollout loss over
$T_\lambda/2$ is kept; training stops after a fixed number of
validation checks without improvement, but not before a minimum number
of updates, and evaluation uses the selected checkpoint. The test windows
are not used for model selection. Window sizes, layer sizes, learning
rates, batch sizes, and stopping counts are listed in
Appendix~\ref{app:algorithm}. Several L63 schedules and two KS runs
reach the update ceiling, and L96 selects its checkpoints during
warm-up (Section~\ref{sec:study:classes}). One run takes
$0.3$--$14$ hours on an A100 shared by two to five runs, for about
$880$ hours of summed process time.

Prediction error is the root-mean-square error over
coordinates, each coordinate normalized by its standard deviation on
the training trajectory (NRMSE; Appendix~\ref{app:algorithm} gives the
formula). Training losses average the same squared normalized errors
over rollout steps and coordinates.
Valid time is the first time this error exceeds $0.3$ (also $0.5$), evaluated at every
model step over the $50$ test windows of $15\,T_\lambda$ and reported in
units of $T_\lambda$; no rollout in the study reached the end of its
window below the threshold, so no valid time is censored. Paired
differences between the adaptive rule and the constant window
at the same budget are
computed seed by seed and reported with $95\%$ confidence intervals from
the $t$ distribution on four degrees of freedom.

Long-time behavior is measured separately on three $200\,T_\lambda$ free
rollouts per run.
Failure is the first non-finite state or a state whose maximum absolute
coordinate exceeds five times the maximum absolute coordinate of the
reference trajectory. We report the failure fraction and first-failure
time, the time at which a long rollout first meets this
criterion.
For a rollout that fails, we compute the MMD, power-spectrum error, and Lyapunov estimate defined below from
the initial part ending $2T_\lambda$ before failure. This usable part
must contain at least $20T_\lambda$ of data.
We compute the marginal maximum mean discrepancy (MMD)
to compare the distributions of state values. We use relative
power-spectrum $L^2$ error to compare how variability is distributed
across frequencies. We also estimate the
largest learned Lyapunov exponent using Benettin's
algorithm~\citep{benettin1980lyapunov} with autodiff Jacobian-vector
products. The exponent is compared with the
true-system estimate from the same algorithm. Statistics from
the usable parts of the rollouts include the number of rollouts or
runs that contributed.

\subsection{Heterogeneity and the coverage gap}
\label{sec:pilot:ftle}

For each system we generated a long post-transient reference trajectory
and computed $\lambda_{\FTLE}(x_0, \tau)$ at $n{=}4000$ sampled points
from the full tangent matrix (Section~\ref{sec:estimators}), sweeping
$\tau \in \{0.5, 1.0, 2.0\}\, T_\lambda$.
\begin{table}[t]
\centering
\scriptsize
\begin{tabular}{lcccccc}
\toprule
System & $\lambda_1$ (ref.) & $\lambda_1$ (data) & $T_\lambda$ & $\CV$ at $\tau{=}T_\lambda/2$ & $\CV$ at $\tau{=}T_\lambda$ & $\CV$ at $\tau{=}2T_\lambda$\\
\midrule
Lorenz-63 & $0.866$ & $0.907$ & $1.155$ & $\mathbf{0.754}$ & $0.559$ & $0.400$ \\
R\"ossler & $0.066$ & -- & $15.152$ & $0.532$ & $0.413$ & $0.423$ \\
Lorenz-96 ($N{=}40$) & $1.733$ & $1.713$ & $0.577$ & $0.106$ & $0.142$ & $0.165$ \\
Kuramoto--Sivashinsky ($N{=}64$, $L{=}22$) & $0.057$ & $0.054$ & $17.544$ & $0.118$ & $0.184$ & $0.227$ \\
\bottomrule
\end{tabular}
\caption{FTLE heterogeneity from $4000$ full-tangent-matrix samples per
system. The working reference $\lambda_1$ defines $T_\lambda$ and all
normalized horizons. The data estimate uses $1000$ time units of
Benettin integration and differs from it by $1$--$5.3\%$; the shared reference
rescales all schedules within a system equally. The listed CVs concern
raw FTLEs; Table~\ref{tab:cor3} reports the floored-rate CV used in
Corollary~\ref{cor:gap}. The heterogeneity ordering does not follow
system dimension.}
\label{tab:hetero}
\end{table}
Table~\ref{tab:hetero} separates the heterogeneous L63 and R\"ossler
systems from L96 and KS. At $\tau=T_\lambda/2$, their CVs span
$0.11$--$0.75$. L63's CV decreases to $0.40$ at $2T_\lambda$,
where it is close to R\"ossler's $0.42$; L96 and KS remain between
$0.11$ and $0.23$ across the measured horizons. Thus the dependence
of heterogeneity on measurement horizon differs across systems.

\label{sec:pilot:coverage}
To evaluate Corollary~\ref{cor:gap}, set
$\beta=\tau\bar\lambda_{\FTLE}$, where $\bar\lambda_{\FTLE}$ is
the mean FTLE over the $4000$ samples, so that the budget equals the
mean stretch over one measurement horizon, and
$\mu_i=\max\{\hat\lambda_{\FTLE,i},L\}$, with $L$ at the $10$th
percentile of positive FTLEs. This floor bounds the longest assigned
window, $W_{\max}=\beta/L$, and handles non-positive rates, observed
in $0.2\%$ of L63 samples at $\tau=T_\lambda/2$ and $2\%$ of
R\"ossler samples at $\tau=T_\lambda$, with none elsewhere in this grid. We compare the
realized gap $R=\bar W^\star/W_g^{\max}-1$, where
$W_g^{\max}=\min_iW_i^\star$, with the lower bound $\CVeff^2$.
The effective CV uses the floored rates, while the raw CV describes
the measured FTLE distribution.
\begin{figure}[t]
\centering
\includegraphics[width=0.85\textwidth]{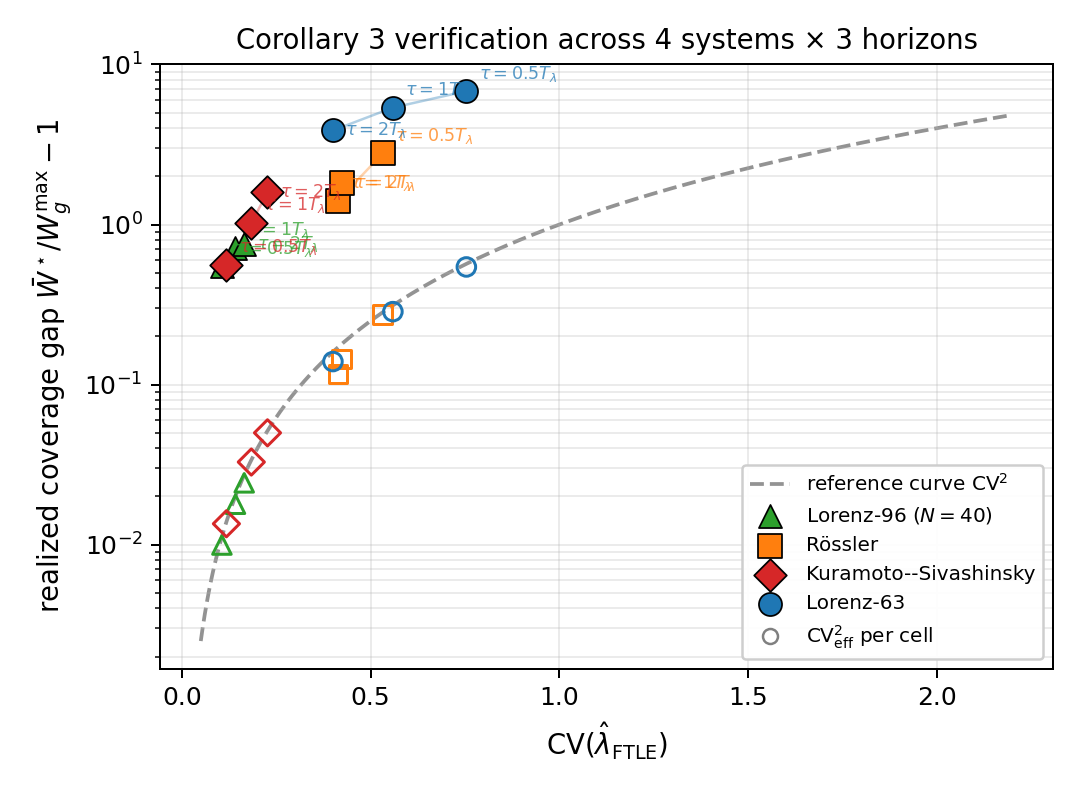}
\caption{Verification of Corollary~\ref{cor:gap}. Realized relative
coverage gap $R = \bar W^\star / W_g^{\max} - 1$ (filled markers) vs.\ the
raw FTLE coefficient of variation $\CV(\hat\lambda_{\FTLE})$ across four
benchmark systems at three horizons each ($\tau \in \{T_\lambda/2,\,
T_\lambda,\, 2T_\lambda\}$, $n{=}4000$ attractor samples per cell, floor at
the $10$th percentile). Hollow markers show the proven lower bound $\CVeff^2$
for the same cell, evaluated at the effective (clipped) CV; the dashed
line is $\CV^2$ as a function of the horizontal axis for reference. Every filled
marker lies above its hollow counterpart. Across the twelve cells the gap
increases with $\CV$ (Pearson $r = 0.90$).}
\label{fig:cor3}
\end{figure}
\begin{table}[t]
\centering
\small
\caption{Numerical check of Corollary~\ref{cor:gap} on four
benchmark systems at three measurement horizons each. Each cell uses
$4000$ sampled states; rates below the $10$th percentile of positive
FTLE values are raised to that floor. $\CV$ is the spread
(coefficient of variation) of the raw FTLE values from
Table~\ref{tab:hetero}, and $\CVeff$ the same spread after flooring.
$W_g^{\max}$ is the longest constant window that the most unstable
sampled state allows, $\bar W^\star$ the mean window of the adaptive
rule at the same budget, and the realized gap is their ratio minus
one. The gap exceeds the proven lower bound $\CVeff^2$ in every cell.
Horizons are in ODE time units.}
\label{tab:cor3}
\begin{tabular}{llcccccc}
\toprule
System & $\tau/T_\lambda$ & $\CV$ & $\CVeff$ & $W_g^{\max}$ & $\bar W^\star$ & realized gap & bound $\CVeff^2$ \\
\midrule
L63 & 0.5 & $0.754$ & $0.737$ & $0.131$ & $1.022$ & $\mathbf{6.82}$ & $0.543$ \\
 & 1.0 & $0.559$ & $0.534$ & $0.230$ & $1.465$ & $5.37$ & $0.286$ \\
 & 2.0 & $0.400$ & $0.373$ & $0.519$ & $2.527$ & $3.87$ & $0.139$ \\
\midrule
R\"ossler & 0.5 & $0.532$ & $0.524$ & $2.438$ & $9.215$ & $2.78$ & $0.275$ \\
 & 1.0 & $0.413$ & $0.341$ & $6.785$ & $16.326$ & $1.41$ & $0.116$ \\
 & 2.0 & $0.423$ & $0.380$ & $12.349$ & $34.814$ & $1.82$ & $0.145$ \\
\midrule
L96 ($N{=}40$) & 0.5 & $0.106$ & $0.100$ & $0.187$ & $0.290$ & $0.55$ & $0.010$ \\
 & 1.0 & $0.142$ & $0.134$ & $0.340$ & $0.584$ & $0.72$ & $0.018$ \\
 & 2.0 & $0.165$ & $0.156$ & $0.669$ & $1.174$ & $0.75$ & $0.024$ \\
\midrule
KS ($L{=}22$) & 0.5 & $0.118$ & $0.116$ & $5.681$ & $8.858$ & $0.56$ & $0.014$ \\
 & 1.0 & $0.184$ & $0.181$ & $8.863$ & $17.949$ & $1.03$ & $0.033$ \\
 & 2.0 & $0.227$ & $0.224$ & $14.098$ & $36.479$ & $1.59$ & $0.050$ \\
\bottomrule
\end{tabular}
\end{table}
The inequality holds in all twelve cells (Table~\ref{tab:cor3}), with
realized gaps $0.55$--$6.8$, or $10$--$55$ times the lower bound.
The adaptive-to-constant coverage ratio reaches $7.8$ on L63 and
$1.5$--$2.6$ on L96 and KS. Across the pooled cells, $R$ correlates
with raw CV ($r=0.90$; $r=0.93$ for $\log R$), but their ordering
is not identical at every horizon. In particular, at $2T_\lambda$
L63 has a larger gap than R\"ossler despite a slightly smaller CV.
The loose bound reflects the dependence of
$R+1=\max_i\mu_i\,\overline{1/\mu}$ on extreme rates.

The floor controls the magnitude of this opportunity
(Table~\ref{tab:clipsweep}). At $\tau=T_\lambda/2$, increasing its
percentile from $1$ to $50$ reduces L63's gap from $9.2$ to $3.3$,
but changes L96 and KS only from $0.56$ to about $0.5$.
The bound and system ordering persist across this floor sweep.
Thus the large gains on heterogeneous systems depend on the maximum
window allowed on low-instability segments; they are not determined
by CV alone. Section~\ref{sec:study:matched} tests the cap's effect
on prediction.

\begin{table}[t]
\centering
\small
\caption{Numerical check of Proposition~\ref{prop:robust} on four
systems, $2000$ sampled states each and $20$ random error draws per
cell. Each rate is perturbed by an error of size up to
$\epsilon/W_{\max}$, so that the accumulated instability over any
allowed window is in error by at most $\epsilon$, and the budget is
shrunk by $\epsilon$, as the proposition prescribes.
Feasibility is the fraction of segments whose true instability stays
within the budget, and the bound ratio is the realized coverage loss
divided by the proposition's bound $2\epsilon\sum_i 1/L_i$; both are
reported over the range of $\epsilon$ tested. The last two columns
repeat the test with Gaussian errors, whose accumulated error can
exceed $\epsilon$ and therefore fall outside the proposition's
assumption. Appendix~\ref{sec:pilot:robust} gives the full protocol.}
\label{tab:prop4}
\begin{tabular}{lcccccc}
\toprule
System & $\beta$ & $\epsilon$ & feasibility (bounded) & bound ratio & Gaussian, $\epsilon \le 0.4$ & Gaussian, $\epsilon \ge 0.8$ \\
\midrule
L63 & $1.43$ & $0.02$--$0.8$ & $100\%$ & $51$--$54\%$ & $99.85$--$99.95\%$ & $100\%$ \\
R\"ossler & $1.36$ & $0.02$--$0.8$ & $100\%$ & $51$--$53\%$ & $99.35$--$99.95\%$ & $100\%$ \\
L96 ($N{=}40$) & $1.62$ & $0.02$--$1.6$ & $100\%$ & $50$--$52\%$ & $98.55$--$99.90\%$ & $100\%$ \\
KS ($L{=}22$) & $2.54$ & $0.02$--$1.6$ & $100\%$ & $52$--$54\%$ & $98.40$--$99.80\%$ & $100\%$ \\
\bottomrule
\end{tabular}
\end{table}

We also check Proposition~\ref{prop:robust} directly
(Table~\ref{tab:prop4}). Starting from the true system's FTLE table on
each benchmark, we add a random error of size up to $\epsilon/W_{\max}$
to every rate, so that the accumulated instability over any allowed
window is in error by at most $\epsilon$, shrink the budget by
$\epsilon$ as the proposition prescribes,
and record whether every segment still respects the true budget and
how much coverage is lost relative to the schedule computed from the
exact rates. Across error sizes from about $1\%$ to over
$50\%$ of the budget, all $26$ cells remain feasible and the coverage
loss is $50$--$54\%$ of the proposition's bound. Gaussian errors,
whose accumulated error can exceed $\epsilon$ and so fall outside the
proposition's assumption, occasionally violate feasibility. These checks confirm the
proposition's arithmetic on linear true-system profiles; they do not
establish the accuracy of any estimator or learned model.
Separate
endpoint-stretch measurements describe true-system growth,
not the learned-model gradient bound. Appendix~\ref{app:diagnostics}
gives that diagnostic, the full robustness protocol, and the
floor-sweep table.

\subsection{Coverage, compute, and accuracy at the same instability budget}
\label{sec:study}
\label{sec:study:matched}

\begin{figure}[t]
\centering
\includegraphics[width=\textwidth]{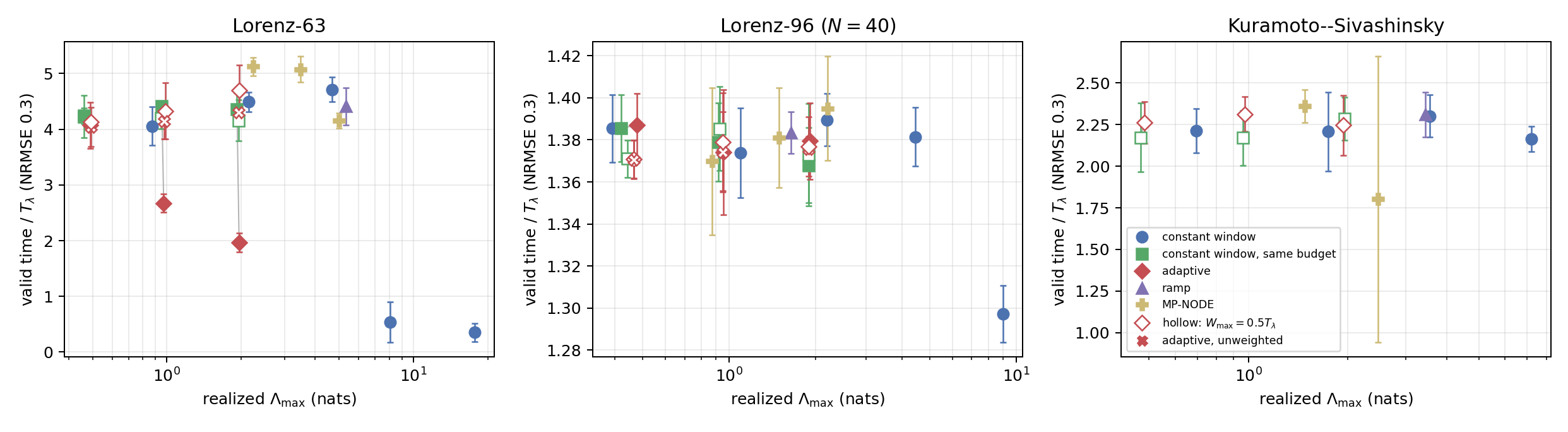}
\caption{Prediction against the budget realized in the final
training batch for every
training schedule (mean $\pm$ standard deviation over five seeds).
Hollow markers use $W_{\max}=\tau=T_\lambda/2$, filled markers
$W_{\max}=2T_\lambda$, and crosses omit inverse-window weighting.
Thin lines connect adaptive and constant schedules at the same budget
and cap. L63 shows poorer prediction for long single rollouts and
large-cap adaptive runs at $\beta\ge1$; its highest means occur for
MP-NODE with $K=3,5$. Similar means elsewhere do not establish
equivalence.}

\label{fig:grid}
\end{figure}

Table~\ref{tab:matched} presents the same-budget comparisons;
Figure~\ref{fig:grid} and Appendix~\ref{app:gridtables} place them in
the full schedule grid. Within each pair, the seed-mean final-batch
budgets agree to within $0.06$ nats. Both schedules use
the same rates, warm-up, and cap. The five-step minimum is the exception to nominal feasibility. On L63
at $\beta=0.5$, it binds on $2$--$3\%$ of segments,
putting $3$--$7\%$ of post-warm-up batches above budget by at most $5.3\%$.
No exceedance is observed at larger budgets or on L96 and KS in the
available post-warm-up audits; L96 contributes only one update per run,
and KS runs stopping before this phase contribute none. During warm-up,
the minimum binds on up to $59\%$ of batches at $\beta=0.5$ and
$9\%$ at $\beta=1$. These batches lie outside the theorem's feasible
set, so the training comparisons concern the same nominal budgets with
this implementation exception.

\begin{table}[t]
\centering
\small
\caption{Adaptive and constant schedules at the same budget.
For each budget $\beta$, we compare the adaptive rule
$W^\star(\beta)$ with the largest constant window feasible for the worst
segment of the batch, $W_g^{\max}(\beta)$, at the same budget and the same
maximum horizon. Valid time is reported in units of $T_\lambda$
as the mean $\pm$ standard deviation across five seeds, using an NRMSE
threshold of $0.3$. The displayed $\Lambda_{\max}$ is the seed mean of the final
training batch's maximum budget; it is not a sensitivity
measurement at the selected checkpoint. $\|\nabla L\|$ is the
seed mean of the final pre-clipping gradient norm. On Lorenz-63 and Lorenz-96
both rules are run with $W_{\max}$ at the FTLE horizon $\tau =
T_\lambda/2$ and extrapolated to $2T_\lambda$, and the capped adaptive
rule also without the inverse-window loss weighting (``unweighted''); on
KS only $W_{\max} = \tau$ fit within the computational limits.}
\label{tab:matched}
\footnotesize
\textbf{Lorenz-63}\\[2pt]
\begin{tabular}{lcccc}
\toprule
$\beta$ & schedule & $\Lambda_{\max}$ & $\|\nabla L\|$ & $t_{\rm valid}/T_\lambda$ \\
\midrule
$0.5$ & $W_g^{\max}(\beta{=}0.5)$, $W_{\max}{=}0.5T_\lambda$ & $0.46$ & $9.2{\rm e}{-5}$ & $4.23 \pm 0.38$ \\
 & $W^\star(\beta{=}0.5)$, $W_{\max}{=}0.5T_\lambda$, inverse-window weighted & $0.49$ & $8.5{\rm e}{-4}$ & $4.13 \pm 0.09$ \\
 & $W^\star(\beta{=}0.5)$, $W_{\max}{=}0.5T_\lambda$, unweighted & $0.49$ & $7.0{\rm e}{-4}$ & $4.05 \pm 0.35$ \\
 & $W_g^{\max}(\beta{=}0.5)$, $W_{\max}{=}2T_\lambda$ & $0.46$ & $1.1{\rm e}{-4}$ & $4.24 \pm 0.14$ \\
 & $W^\star(\beta{=}0.5)$, $W_{\max}{=}2T_\lambda$, inverse-window weighted & $0.49$ & $6.9{\rm e}{-4}$ & $4.07 \pm 0.41$ \\
\midrule
$1$ & $W_g^{\max}(\beta{=}1)$, $W_{\max}{=}0.5T_\lambda$ & $0.95$ & $3.3{\rm e}{-4}$ & $4.20 \pm 0.19$ \\
 & $W^\star(\beta{=}1)$, $W_{\max}{=}0.5T_\lambda$, inverse-window weighted & $0.99$ & $2.2{\rm e}{-3}$ & $4.33 \pm 0.51$ \\
 & $W^\star(\beta{=}1)$, $W_{\max}{=}0.5T_\lambda$, unweighted & $0.98$ & $2.2{\rm e}{-3}$ & $4.15 \pm 0.32$ \\
 & $W_g^{\max}(\beta{=}1)$, $W_{\max}{=}2T_\lambda$ & $0.95$ & $3.6{\rm e}{-4}$ & $4.41 \pm 0.09$ \\
 & $W^\star(\beta{=}1)$, $W_{\max}{=}2T_\lambda$, inverse-window weighted & $0.97$ & $6.5{\rm e}{-2}$ & $2.67 \pm 0.16$ \\
\midrule
$2$ & $W_g^{\max}(\beta{=}2)$, $W_{\max}{=}0.5T_\lambda$ & $1.96$ & $7.1{\rm e}{-4}$ & $4.16 \pm 0.37$ \\
 & $W^\star(\beta{=}2)$, $W_{\max}{=}0.5T_\lambda$, inverse-window weighted & $1.96$ & $5.0{\rm e}{-3}$ & $4.70 \pm 0.45$ \\
 & $W^\star(\beta{=}2)$, $W_{\max}{=}0.5T_\lambda$, unweighted & $1.95$ & $4.1{\rm e}{-3}$ & $4.30 \pm 0.15$ \\
 & $W_g^{\max}(\beta{=}2)$, $W_{\max}{=}2T_\lambda$ & $1.93$ & $1.1{\rm e}{-3}$ & $4.36 \pm 0.30$ \\
 & $W^\star(\beta{=}2)$, $W_{\max}{=}2T_\lambda$, inverse-window weighted & $1.96$ & $9.8{\rm e}{-1}$ & $1.96 \pm 0.17$ \\
\bottomrule
\end{tabular}
\\[6pt]
\textbf{Kuramoto--Sivashinsky}\\[2pt]
\begin{tabular}{lcccc}
\toprule
$\beta$ & schedule & $\Lambda_{\max}$ & $\|\nabla L\|$ & $t_{\rm valid}/T_\lambda$ \\
\midrule
$0.5$ & $W_g^{\max}(\beta{=}0.5)$, $W_{\max}{=}0.5T_\lambda$ & $0.47$ & $2.5{\rm e}{-3}$ & $2.17 \pm 0.21$ \\
 & $W^\star(\beta{=}0.5)$, $W_{\max}{=}0.5T_\lambda$, inverse-window weighted & $0.48$ & $9.4{\rm e}{-3}$ & $2.26 \pm 0.12$ \\
\midrule
$1$ & $W_g^{\max}(\beta{=}1)$, $W_{\max}{=}0.5T_\lambda$ & $0.96$ & $1.8{\rm e}{-2}$ & $2.17 \pm 0.17$ \\
 & $W^\star(\beta{=}1)$, $W_{\max}{=}0.5T_\lambda$, inverse-window weighted & $0.97$ & $4.7{\rm e}{-2}$ & $2.31 \pm 0.10$ \\
\midrule
$2$ & $W_g^{\max}(\beta{=}2)$, $W_{\max}{=}0.5T_\lambda$ & $1.96$ & $4.5{\rm e}{-2}$ & $2.29 \pm 0.13$ \\
 & $W^\star(\beta{=}2)$, $W_{\max}{=}0.5T_\lambda$, inverse-window weighted & $1.94$ & $1.3{\rm e}{-1}$ & $2.25 \pm 0.18$ \\
\bottomrule
\end{tabular}
\\[6pt]
\textbf{Lorenz-96 ($N{=}40$)}\\[2pt]
\begin{tabular}{lcccc}
\toprule
$\beta$ & schedule & $\Lambda_{\max}$ & $\|\nabla L\|$ & $t_{\rm valid}/T_\lambda$ \\
\midrule
$0.5$ & $W_g^{\max}(\beta{=}0.5)$, $W_{\max}{=}0.5T_\lambda$ & $0.44$ & $2.4{\rm e}{-3}$ & $1.37 \pm 0.01$ \\
 & $W^\star(\beta{=}0.5)$, $W_{\max}{=}0.5T_\lambda$, inverse-window weighted & $0.47$ & $3.5{\rm e}{-3}$ & $1.37 \pm 0.01$ \\
 & $W^\star(\beta{=}0.5)$, $W_{\max}{=}0.5T_\lambda$, unweighted & $0.47$ & $3.6{\rm e}{-3}$ & $1.37 \pm 0.01$ \\
 & $W_g^{\max}(\beta{=}0.5)$, $W_{\max}{=}2T_\lambda$ & $0.42$ & $2.6{\rm e}{-3}$ & $1.39 \pm 0.02$ \\
 & $W^\star(\beta{=}0.5)$, $W_{\max}{=}2T_\lambda$, inverse-window weighted & $0.48$ & $4.9{\rm e}{-3}$ & $1.39 \pm 0.02$ \\
\midrule
$1$ & $W_g^{\max}(\beta{=}1)$, $W_{\max}{=}0.5T_\lambda$ & $0.93$ & $1.0{\rm e}{-2}$ & $1.39 \pm 0.02$ \\
 & $W^\star(\beta{=}1)$, $W_{\max}{=}0.5T_\lambda$, inverse-window weighted & $0.95$ & $1.6{\rm e}{-2}$ & $1.38 \pm 0.02$ \\
 & $W^\star(\beta{=}1)$, $W_{\max}{=}0.5T_\lambda$, unweighted & $0.95$ & $1.6{\rm e}{-2}$ & $1.37 \pm 0.02$ \\
 & $W_g^{\max}(\beta{=}1)$, $W_{\max}{=}2T_\lambda$ & $0.92$ & $9.7{\rm e}{-3}$ & $1.38 \pm 0.02$ \\
 & $W^\star(\beta{=}1)$, $W_{\max}{=}2T_\lambda$, inverse-window weighted & $0.96$ & $1.5{\rm e}{-2}$ & $1.37 \pm 0.03$ \\
\midrule
$2$ & $W_g^{\max}(\beta{=}2)$, $W_{\max}{=}0.5T_\lambda$ & $1.89$ & $4.7{\rm e}{-2}$ & $1.37 \pm 0.02$ \\
 & $W^\star(\beta{=}2)$, $W_{\max}{=}0.5T_\lambda$, inverse-window weighted & $1.89$ & $5.8{\rm e}{-2}$ & $1.38 \pm 0.01$ \\
 & $W^\star(\beta{=}2)$, $W_{\max}{=}0.5T_\lambda$, unweighted & $1.89$ & $5.7{\rm e}{-2}$ & $1.38 \pm 0.01$ \\
 & $W_g^{\max}(\beta{=}2)$, $W_{\max}{=}2T_\lambda$ & $1.89$ & $3.4{\rm e}{-2}$ & $1.37 \pm 0.02$ \\
 & $W^\star(\beta{=}2)$, $W_{\max}{=}2T_\lambda$, inverse-window weighted & $1.91$ & $5.6{\rm e}{-2}$ & $1.38 \pm 0.02$ \\
\bottomrule
\end{tabular}

\end{table}

With $W_{\max}=\tau$, adaptive-minus-constant valid-time differences
on L63 are $-0.10$, $+0.13$, and $+0.54T_\lambda$ at
$\beta=0.5,1,2$, with paired $95\%$ confidence intervals
$[-0.69,0.49]$, $[-0.52,0.78]$, and $[-0.05,1.13]$.
The KS differences are $+0.09$, $+0.14$, and $-0.04T_\lambda$,
with interval half-widths $0.2$--$0.35$; every L96 pair differs by
at most $0.01T_\lambda$, with half-width at most $0.05$.
The unweighted L63 rule reaches $4.05$, $4.15$, and $4.30T_\lambda$,
within $0.2T_\lambda$ of its constant counterpart.
The smaller-cap comparisons establish no consistent advantage in five
seeds. Their intervals include zero but also permit meaningful gains,
up to $1.13T_\lambda$ on L63 and $0.46T_\lambda$ on KS.
Equivalence would require a prespecified practical margin and enough
precision to assess it.

The coverage gain is large. At $W_{\max}=\tau$, adaptive-to-constant
ratios of seed-mean final-batch horizons are $4.3$, $3.3$, and $2.1$
on L63 at $\beta=0.5,1,2$, about $1.3$ on KS, and $1.1$--$1.2$
on L96. These implemented ratios reflect the grouping of windows by
length (Section~\ref{sec:bootstrap}), discretization,
and minimum-window exceptions and are not predicted by the CV bound
alone. The nominal ungrouped L63 schedule would exceed the cap on
$20\%$, $43\%$, and $76\%$ of sampled states; on L96 at $\beta=2$,
$47\%$ of logged segments reach the cap, while no logged KS adaptive
segment does so. Collecting this coverage requires $2.3$--$3.9$ times as many model
steps per optimizer update on L63 and $1.2$--$1.3$ times on L96 and
KS. Wall-clock cost per update rises by $3.6$--$4.7$ times on L63
and KS and $2$--$4$ times on L96, because unequal horizons require
separate rollout groups. Thus the coverage gains incur additional work
without a clear improvement in prediction accuracy under this training protocol.

Increasing the L63 cap to $2T_\lambda$ changes the outcome. At
$\beta=1,2$, adaptive-minus-constant valid-time differences are
$-1.74T_\lambda$ $[-2.00,-1.48]$ and
$-2.40T_\lambda$ $[-2.68,-2.11]$, with degradation in every seed.
Final gradient norms are $0.07$ and $1.0$, versus
$4\times10^{-4}$ and $10^{-3}$ for the constant counterparts, despite
nominal realized budgets of $0.97$ and $1.96$ nats. These gradients
also depend on losses, parameters, and sensitivity prefactors and
therefore do not measure the true instability budget. To investigate a possible role for proxy error, we compare
the FTLE tables of Section~\ref{sec:pilot:ftle}, whose $4000$ sample
states are the same at every $\tau$. On Lorenz-63 the rate of a
segment over $\tau = T_\lambda/2$ has correlation $0.26$ with its rate
over $2T_\lambda$. The decile of segments with the lowest rate over
$T_\lambda/2$, mean $0.29$, has a rate over $2T_\lambda$ of $1.05$, so a window
of $2T_\lambda$ on those states has mean true-system endpoint stretch
$2.41$ nats, compared with the short-rate proxy's mean $0.66$ nats. The adaptive rule
assigns exactly those segments its longest windows. On the nominal schedule computed on the same states, before
grouping windows by length and separate from the window lengths recorded
during training, the rule sends $43\%$ of states beyond $\tau$ and $9\%$
to the $2T_\lambda$ cap at $\beta = 1$, and $76\%$ and $20\%$ at
$\beta = 2$.

A complementary diagnostic evaluates rates over horizons
covering the assigned windows. Choose the smallest saved horizon
$\tau^*\in\{T_\lambda/2,T_\lambda,2T_\lambda\}$ covering the assigned
window and compute $W_i\widehat\lambda_i(\tau^*)$.
For $\beta=1$ and $2$, this diagnostic has $95$th percentiles $2.71$
and $3.30$ nats and maxima above $8$. At the exact $2T_\lambda$ cap,
the mean true-system endpoint stretches are $2.43$ and $2.49$ nats,
compared with nominal means $0.60$ and $1.12$.
For intermediate windows, multiplying a longer-horizon rate by $W_i$
remains a proxy; it is not an exact endpoint measurement at $W_i$.
At $\beta=0.5$, only $3\%$ of states reach the cap and the corresponding
training comparison shows no clear degradation ($4.07$ versus $4.24$).
These discrepancies do not supply the learned-flow uniform error bound
required by Proposition~\ref{prop:robust}; its sufficient condition
neither certifies feasibility here nor predicts failure when unverified.

Long constant L63 windows also perform poorly: $W=1,2T_\lambda$
yield $0.54\pm0.36$ and $0.35\pm0.16T_\lambda$ at final
budgets of about $8$ and $18$ nats. On L96, large-cap
adaptive configurations reach $1.37$--$1.39T_\lambda$. Their nominal
windows, computed on the $4000$-sample FTLE tables of
Section~\ref{sec:pilot:ftle}, do not reach the $2T_\lambda$ cap, and the mean
long-horizon true-system rate is below the short-horizon mean. This does
not certify conservative error for every state or explain the training
outcome, which also reflects early checkpoint selection. The cap is the only configuration change between the two adaptive
versions: budgets, warm-up, window grouping, and seeds are shared. Reducing it
removes the observed L63 degradation at the larger budgets, but also
changes window lengths, loss weighting, compute, and sampled starts.
The experiment therefore supports the smaller cap for this setup
without isolating profile extrapolation as the cause or establishing
a general safety condition. The CV across starting states does not, by itself, show
whether the estimates remain accurate over longer windows or whether
training will fail.

\subsection{Broader schedule comparisons}
\label{sec:study:classes}

The wider grid shows effects that the paired adaptive comparisons do
not resolve. On L63, constant windows of $0.1$--$0.5T_\lambda$
reach $4.05$--$4.72T_\lambda$, the six same-budget constant windows reach
$4.16$--$4.41T_\lambda$, and the ramp reaches $4.41\pm0.33T_\lambda$.
The MP-NODE schedules reach $4.15\pm0.14$, $5.08\pm0.24$, and
$5.13\pm0.16T_\lambda$ at $K=2,3,5$. The latter two have the highest
means, about $0.4T_\lambda$ above the best single-window result,
at six times those constant windows' cost per update. The $K=2,3$
runs and the $0.5T_\lambda$ constant window reach the update ceiling in
every seed; $K=5$ does so in three of five. Their validation-selected
performance could change with further training.

Two exploratory paired comparisons illustrate this variation. At the
same nominal total window and approximately equal forward work,
$K=3$ exceeds $K=2$ by $0.92T_\lambda$ $[0.54,1.30]$.
The $0.5T_\lambda$ constant window exceeds $0.1T_\lambda$ by
$0.66T_\lambda$ $[0.11,1.21]$. These nominal intervals, together
with the poor long-window results, preclude a general conclusion that
rollout allocation is unimportant. Most KS configuration means lie near $2.2$--$2.4T_\lambda$.
The $1T_\lambda$ constant window reaches $2.16\pm0.08T_\lambda$
with a final budget of $7$ nats and gradient norm $3$, taking
$9$--$14$ hours per seed. MP-NODE with $K=5$ reaches
$2.36\pm0.10T_\lambda$, while $K=3$ reaches $1.80\pm0.86T_\lambda$,
including one seed at $0.10T_\lambda$. These means and seed variation
do not establish equality across schedules.

On L96, all schedules reach $1.37$--$1.39T_\lambda$ except the
$2T_\lambda$ constant window at $1.30\pm0.01T_\lambda$.
All $120$ runs select checkpoints at steps $600$--$4{,}000$ and stop
at the $15{,}000$-step minimum (Appendix~\ref{app:algorithm}). Budgeted schedules are therefore
evaluated at $34$--$66\%$ of their target budgets, during a warm-up
starting at $25\%$, rather than after the schedules reach their target budgets.
Appendix~\ref{app:gridtables} gives every configuration's results at
both prediction thresholds. The per-segment gradient diagnostics in
Appendix~\ref{sec:study:lemma} are descriptive: their horizontal axis is the
true-system FTLE proxy, which does not test the learned-model bound in Lemma~\ref{lem:sensitivity}.

\subsection{Long-time behavior beyond short-time prediction}
\label{sec:study:longtime}

\begin{table}[t]
\centering
\small
\caption{Long-rollout behavior of every trained model. Each run
integrates three $200\,T_\lambda$ rollouts from different starting
states. A rollout fails when it becomes non-finite or exceeds five
times the largest reference amplitude. Entries show ranges across
schedule means. Failure summaries use all five seeds. The time to first
failure is the mean time at which a rollout fails; $200$ indicates that
no rollout failed during evaluation. The learned Lyapunov exponent
error, marginal MMD (a distance between the learned and reference state
distributions), and power-spectrum error are computed from complete
rollouts when no failure occurs and usable initial parts when a rollout
fails. For KS, only runs with a sufficiently long usable part contribute
to these three statistics. The contributing seed counts are given in
Appendix~\ref{app:gridtables}. The two long constant windows on
Lorenz-63 are listed separately because their rollouts stay bounded
but settle on a different attractor from the true one.}
\label{tab:longtime}
\scriptsize
\setlength{\tabcolsep}{4pt}
\begin{tabular}{lcccccc}
\toprule
System & \shortstack{schedules\\(runs)} & \shortstack{rollouts\\reaching $200T_\lambda$} & \shortstack{time to first\\failure ($T_\lambda$)} & $|\hat\lambda_1-\lambda_1|$ & \shortstack{marginal MMD\\($\times10^{-3}$)} & \shortstack{power-spectrum\\error} \\
\midrule
Lorenz-63 (valid time $>1T_\lambda$) & $22$ ($110$) & all & $200$ (none) & $0.011$--$0.023$ & $0.5$--$2.9$ & $0.16$--$0.25$ \\
Lorenz-63 ($1$, $2T_\lambda$ windows) & $2$ ($10$) & all & $200$ (none) & $1.24$--$1.62$ & $196$--$376$ & $0.72$--$0.99$ \\
Lorenz-96 ($N{=}40$) & $24$ ($120$) & all & $200$ (none) & $0.068$--$0.222$ & $1.7$--$2.5$ & $0.08$--$0.12$ \\
Kuramoto--Sivashinsky & $13$ ($65$) & none & $13$--$28$ & $0.006$--$0.021$ & $1.3$--$11.3$ & $0.20$--$0.66$ \\
\bottomrule
\end{tabular}

\end{table}

Table~\ref{tab:longtime} summarizes the long-rollout
statistics. On Lorenz-63 and Lorenz-96 every run with a valid time above
$1\,T_\lambda$ produces $200\,T_\lambda$ rollouts that remain finite and
bounded under the evaluation criterion (five times the maximum
absolute reference amplitude). Averaged over
seeds per schedule, the learned Lyapunov exponent is within $0.01$ to
$0.023$ of the true $0.907$ on Lorenz-63 and within $0.07$ to $0.14$ of
the true $1.71$ on Lorenz-96 ($0.22$ for the $2\,T_\lambda$ window), the
marginal MMD is $0.5$ to $3 \times 10^{-3}$, and the power-spectrum
error is $0.16$ to $0.25$ on Lorenz-63 and $0.08$ to $0.12$ on
Lorenz-96. These configuration means are descriptive and do not
establish equivalence of long-time statistics across schedules.
The Lorenz-63 models trained with the $1$ and $2T_\lambda$
constant windows predict poorly but remain bounded. They settle on
a different attractor from the true one, with
power-spectrum errors of $0.7$ to $1.0$, MMDs of $0.2$ to $0.4$ and
Lyapunov errors of $1.2$ to $1.6$.

On KS, most configurations have mean valid times near two
Lyapunov times, but all evaluated models eventually cross the
long-rollout failure criterion: none of the $195$ independently integrated rollouts survives
the full $200\,T_\lambda$. Their first-failure times run from $3$ to
$66\,T_\lambda$ with median $16$ and mean $19$, and the schedule means
range from $13$ (MP-NODE, $K{=}3$) to $28\,T_\lambda$ (the
$0.1\,T_\lambda$ window), with seed-to-seed spreads as large as the
differences between schedules (Table~\ref{tab:grid_ks}). Within a run
the three rollouts, started from different states, cross the failure criterion
within a median range of $1.6T_\lambda$ of one another. This
describes variation over the three tested starts per run and does not
establish independence from the starting state. For KS, the Lyapunov error, marginal MMD, and power-spectrum error
are computed only from rollouts with sufficiently long usable data
before failure. These results therefore describe a subset of the
evaluated rollouts. For this subset, the schedule means give Lyapunov
errors of $0.006$ to $0.021$ for a true exponent of $0.054$,
marginal MMDs of $1.3$
to $11\times10^{-3}$, and power-spectrum errors of $0.20$ to $0.66$. For KS, we integrated each rollout separately so that a failure
in one rollout did not stop the others. The schedule-mean failure horizons should be interpreted
alongside their seed variation, not as evidence that schedule effects
are absent.

\section{Discussion and limitations}
\label{sec:pilot:summary}
\label{sec:practical}

\subsection{Main findings}
The theory shows how to use more trajectory data while staying
within a chosen instability limit. The training study achieves this
greater coverage at additional cost, but whether it improves prediction
remains uncertain. The paired confidence intervals
still allow meaningful gains or losses in prediction accuracy, and comparisons with other
schedules show differences in accuracy. The paired
comparisons therefore do not show that the schedules give equal
prediction accuracy or that the choice of window length is unimportant.

Variation in instability across starting states does not indicate
whether a rate measured over a short window also describes a longer
one. The rate measurements and cap experiment help separate these
questions. In the
$4000$-sample FTLE tables of Section~\ref{sec:pilot:ftle},
short/long-horizon rate correlations are $0.264$ on L63, $0.356$ on
L96, and $0.042$ on KS despite KS's low short-horizon CV.
Neither CV nor the correlation alone indicates whether an
instability estimate is too small and will harm training. The smaller
cap improves prediction in the tested L63 configurations, but also
changes other parts of training, so the experiment does not establish
which change explains the improvement. Instability estimates should
therefore be checked over the window lengths actually used.

The KS results show that useful short-time prediction can
coexist with failure over longer times. The reported failure times
apply to the tested models, starting states, and definition of failure.
Evaluating a rollout schedule therefore requires measuring coverage,
computational cost, and prediction accuracy for the model selected by
validation, as well as testing long-time behavior when it matters.

\subsection{Limitations}
\label{sec:limitations}
The study covers one model per system, one FTLE horizon
($\tau=T_\lambda/2$), one budget warm-up, five seeds, and nominal
budgets up to $2$ nats. It does not test training with profiles measured
over each assigned window or with a running tangent integral.
The use of the true system's FTLE as a proxy and minimum-window exceptions prevent interpreting
the implemented budgets as certified learned-model sensitivity bounds.
The available post-warm-up audits are especially sparse on L96 and KS
(Section~\ref{sec:study:matched}).

Early checkpoint selection on L96 and runs that reach the
update ceiling on L63 limit
inferences about fully developed budgets and longer training
(Section~\ref{sec:study:classes}). Our MP-NODE baseline uses a
fixed penalty in our own implementation, rather than reference code
or a search for the published method's best performance. Costs follow
the stated stopping protocol, so the study does not rank accuracy at
fixed compute. Memory constraints exclude KS windows beyond
$1T_\lambda$ and the large-cap adaptive rules; the KS statistics
computed before failure represent only runs with enough usable data.

Comparisons at equal computational cost and with less training data would test when
extra coverage has greater value. Profiles evaluated along the learned
flow could assess the gap between the true-system proxy and the
learned model, and the L63 MP-NODE
results motivate extending the allocation framework. Long-time KS
stability, model-based proxies, noise, partial observation, and
statistical objectives remain open; weak-penalty methods offer a related
direction for noisy training \citep{li2025weak}.

\subsection{Conclusion}
The analysis shows how to cover the most trajectory data within a
chosen instability budget. In the $305$-run training study, adapting
the window length to each starting state increases coverage and
computational cost without showing a consistent improvement in
prediction over constant windows
at the same instability budget, computed from the true system's
FTLE, and the same horizon cap. On Lorenz-63, a smaller cap
removes the drop in prediction accuracy seen when windows extend beyond
the FTLE measurement horizon, although the reason for the improvement
remains unresolved. The KS failures show that useful short-time
prediction can coexist with poor long-time behavior. These findings
support measuring coverage, computational cost, and prediction accuracy
together, while reporting uncertainty in the results. They also support
checking instability estimates over the windows used in training and
testing long-time rollouts separately.

\section*{Reproducibility}
The code for this study is available at
\url{https://github.com/xupingxie/node-rollout}. The repository contains
the dataset-generation, training, and post-processing scripts and the
configuration files; the datasets are generated by these scripts. Section~\ref{sec:setup} and
Appendix~\ref{app:algorithm} list the data generation, model, and
training settings, and every table reports the number of runs or seeds
behind each statistic.

\makeatletter\if@accepted
\section*{Acknowledgments}
This research used resources of the National Energy Research Scientific
Computing Center, a DOE Office of Science User Facility supported by the
Office of Science of the U.S. Department of Energy under Contract
No.~DE-AC02-05CH11231 using NERSC award NERSC DDR-ERCAP0035032.
\fi\makeatother

\bibliographystyle{tmlr}
\bibliography{references}

\clearpage
\appendix

\section{Detailed proofs}
\label{app:proof}

\subsection{Proof of Lemma~\ref{lem:sensitivity}}

Differentiating~\eqref{eq:segloss} with respect to $\theta$ gives
\[
\nabla_\theta \mathcal{L}^{(i)}_{\mathrm{roll}}
= \sum_j \omega_{i,j}\, \partial_x \ell\bigl(x_\theta(t_{i,j}), x^\star(t_{i,j})\bigr)\, S_i(t_{i,j}).
\]
Using \eqref{eq:varofconst} and Assumption~\ref{assump:bounds},
\[
\|S_i(t_{i,j})\|
\le C_\Phi B_\theta \int_0^{t_{i,j}} \exp\!\left(\int_s^{t_{i,j}} \mu_i(r)\,dr\right) ds.
\]
Substitution and the bound $\|\partial_x \ell\| \le B_\ell$ yield~\eqref{eq:gradbound}. Bounding the
inner exponential by $e^\beta$ when $\Lambda_i(W_i) \le \beta$
gives~\eqref{eq:gradbudget}. \hfill$\square$

\subsection{Proof of Theorem~\ref{thm:optimality}}

\begin{proof}[Detailed proof]
Recall $\Lambda_i: [W_{\min}, W_{\max}] \to \R_{\ge 0}$ is non-decreasing
and continuous, hence its level sets are closed intervals. Define
$W_i^\star(\beta) = \max\{ W \in [W_{\min}, W_{\max}] : \Lambda_i(W) \le \beta\}$.
Existence of the maximum follows from continuity and compactness of the
domain.

\emph{(i) Feasibility of $W^\star(\beta)$.}
$\Lambda_i(W_i^\star(\beta)) \le \beta$ by construction, so
$G(W^\star(\beta)) = \max_i \Lambda_i(W_i^\star(\beta)) \le \beta$, hence
$W^\star(\beta) \in \F(\beta)$.

\emph{(ii) Pointwise optimality.}
Let $W \in \F(\beta)$ be arbitrary. Then $\Lambda_i(W_i) \le \beta$ for each
$i$, so by definition of the saturating horizon $W_i \le W_i^\star(\beta)$
for each $i$. Summing,
\[
C(W) = \sum_{i=1}^B W_i \le \sum_{i=1}^B W_i^\star(\beta) = C(W^\star(\beta)).
\]
Hence $W^\star(\beta)$ attains the supremum of $C$ on $\F(\beta)$.

Uniqueness needs no further assumption: if some $W \in \F(\beta)$ satisfies
$C(W) = C(W^\star(\beta))$, then $\sum_i (W_i^\star - W_i) = 0$ together
with $W_i \le W_i^\star$ for every $i$ forces $W_i = W_i^\star$ for all $i$.

\emph{(iii) Strict dominance over global rules.}
A global schedule $W^g = (W_g, \ldots, W_g)$ lies in $\F(\beta)$ iff
$\Lambda_i(W_g) \le \beta$ for all $i$, equivalently $W_g \le W_i^\star(\beta)$
for all $i$, equivalently $W_g \le \min_i W_i^\star(\beta)$. Thus
$C(W^\star) - C(W^g) = \sum_i (W_i^\star - W_g) \ge 0$, a sum of
non-negative terms, with equality iff $W_i^\star = W_g$ for all $i$. For
$W_g = \min_i W_i^\star$ this says: zero gap iff all $W_i^\star$ coincide.
\end{proof}

\begin{remark}[When the gap vanishes]
The gap in Theorem~\ref{thm:optimality}(iii) against the largest feasible
global window is zero exactly when all saturating horizons coincide,
which happens when the profiles cross the level $\beta$ at a common
horizon or are all clipped at $W_{\max}$. On the attractors of
Section~\ref{sec:experiments} the saturating horizons are spread and the
gap is positive; Corollary~\ref{cor:gap} bounds it from below.
\end{remark}

\subsection{Proof of Corollary~\ref{cor:gap}}
\label{app:cor-proof}

\begin{proof}
The identity~\eqref{eq:gapratio} is immediate from
$C(W^\star) = B\bar W^\star$ and $C(W_g^{\max}\mathbf 1) = B W_g^{\max}$.
Non-negativity and the equality case follow from
Theorem~\ref{thm:optimality}(iii).

For the bound~\eqref{eq:cvbound}, assume $\Lambda_i(W) = \mu_i W$ with
$\mu_i > 0$ and that $\beta/\mu_i \in [W_{\min}, W_{\max}]$ for every
$i$, so that $W_i^\star(\beta) = \beta/\mu_i$ and
$W_g^{\max} = \min_i \beta/\mu_i = \beta/\max_i \mu_i$. Write
$\bar\mu = \tfrac1B\sum_i \mu_i$, $\sigma^2 = \tfrac1B\sum_i \mu_i^2 -
\bar\mu^2$ and $\CVeff = \sigma/\bar\mu$. Then
\[
R + 1 \;=\; \frac{\bar W^\star}{W_g^{\max}}
\;=\; \Bigl(\frac1B\sum_{i=1}^B \frac{1}{\mu_i}\Bigr)\,\max_i \mu_i .
\]
The map $x \mapsto 1/x$ is convex on $(0,\infty)$, so by Jensen's
inequality $\tfrac1B\sum_i \mu_i^{-1} \ge 1/\bar\mu$. Since each
$\mu_i \ge 0$, $\mu_i^2 \le \mu_i \max_j \mu_j$, and averaging gives
$\tfrac1B\sum_i \mu_i^2 \le \bar\mu \max_j \mu_j$, i.e.
\[
\max_j \mu_j \;\ge\; \frac{\tfrac1B\sum_i \mu_i^2}{\bar\mu}
\;=\; \frac{\bar\mu^2 + \sigma^2}{\bar\mu}
\;=\; \bar\mu\,(1 + \CVeff^2).
\]
Multiplying the two inequalities,
$R + 1 \ge \bar\mu^{-1}\cdot \bar\mu (1 + \CVeff^2) = 1 + \CVeff^2$,
which is~\eqref{eq:cvbound}. Both inequalities are equalities if and
only if all $\mu_i$ are equal, in which case $R = \CVeff = 0$; away from
that case the bound is loose (two rates $1$ and $2$ with equal mass give
$R = 1/2$ against $\CVeff^2 = 1/9$), and the experiments find it loose
by an order of magnitude.
\end{proof}

\begin{remark}[A stronger CV bound need not hold]
The alternative bound $R\ge\CV/(1+\CV)$ fails even for a simple
finite batch: $99$ segments at $\mu=1$ and one at $\mu=0.5$ give
$R=0.010$ but $\CV/(1+\CV)=0.048$.
\end{remark}

\subsection{Proof of Proposition~\ref{prop:robust}}
\label{app:robust-proof}

\begin{proof}
Fix $\beta \ge 0$ and $\epsilon \ge 0$ with $\beta \ge \epsilon +
\widehat\Lambda_i(W_{\min})$ for all $i$. By continuity and monotonicity
of $\widehat\Lambda_i$, the set $\{W \in [W_{\min}, W_{\max}] :
\widehat\Lambda_i(W) \le \beta - \epsilon\}$ is a nonempty closed
interval containing $W_{\min}$, so its supremum $\widetilde W_i$ is
attained and $\widehat\Lambda_i(\widetilde W_i) \le \beta - \epsilon$.

\emph{(i) Feasibility.} For each $i$, the uniform error
bound~\eqref{eq:eps-bound} gives
\[
\Lambda_i(\widetilde W_i)
\;\le\; \widehat\Lambda_i(\widetilde W_i) + \epsilon
\;\le\; (\beta - \epsilon) + \epsilon \;=\; \beta.
\]
Hence $G(\widetilde W) = \max_i \Lambda_i(\widetilde W_i) \le \beta$, so
$\widetilde W \in \F(\beta)$.

\emph{(ii) Coverage gap.} Fix $i$ and consider two cases.

\emph{Case 1: $\widetilde W_i = W_{\max}$.} Then $\widetilde W_i \le
W_i^\star(\beta) \le W_{\max} = \widetilde W_i$ (the lower inequality
follows from feasibility, since $\Lambda_i(\widetilde W_i) \le \beta$
implies $\widetilde W_i$ is in the feasible set defining
$W_i^\star(\beta)$). So $W_i^\star(\beta) - \widetilde W_i = 0$.

\emph{Case 2: $\widetilde W_i < W_{\max}$.} Then the constraint
in~\eqref{eq:tildeW} is binding at $\widetilde W_i$, so by continuity of
$\widehat\Lambda_i$,
\begin{equation}
\widehat\Lambda_i(\widetilde W_i) \;=\; \beta - \epsilon.
\label{eq:binding}
\end{equation}
The error bound~\eqref{eq:eps-bound} then gives
\begin{equation}
\Lambda_i(\widetilde W_i) \;\ge\; \widehat\Lambda_i(\widetilde W_i) - \epsilon
\;=\; \beta - 2\epsilon.
\label{eq:lower-true}
\end{equation}
On the other hand, $\Lambda_i(W_i^\star(\beta)) \le \beta$ by the
definition of $W_i^\star(\beta)$. Feasibility (part (i)) gives
$\widetilde W_i \le W_i^\star(\beta)$, so subtracting~\eqref{eq:lower-true}
from $\Lambda_i(W_i^\star(\beta)) \le \beta$ yields
\begin{equation}
\Lambda_i(W_i^\star(\beta)) - \Lambda_i(\widetilde W_i)
\;\le\; \beta - (\beta - 2\epsilon) \;=\; 2\epsilon.
\label{eq:lambda-gap}
\end{equation}
The strong-monotonicity hypothesis $\mu_i(t) \ge L_i$ a.e.\ gives
\[
\Lambda_i(W_i^\star(\beta)) - \Lambda_i(\widetilde W_i)
\;=\; \int_{\widetilde W_i}^{W_i^\star(\beta)} \mu_i(r)\, dr
\;\ge\; L_i \bigl(W_i^\star(\beta) - \widetilde W_i\bigr).
\]
Combining with~\eqref{eq:lambda-gap}, $L_i (W_i^\star(\beta) -
\widetilde W_i) \le 2\epsilon$, i.e.,
\[
W_i^\star(\beta) - \widetilde W_i \;\le\; \frac{2\epsilon}{L_i},
\]
which is~\eqref{eq:per-seg-gap}. Summing over $i = 1, \ldots, B$ and
combining the two cases yields the total-coverage
bound~\eqref{eq:total-gap}.
\end{proof}

\begin{remark}[Two-sided perturbation interpretation]
\label{rem:two-sided}
The factor $2$ in~\eqref{eq:per-seg-gap} combines budget shrinkage
by $\epsilon$ with the lower bound
$\Lambda_i(\widetilde W_i)\ge\widehat\Lambda_i(\widetilde W_i)-\epsilon$
at the estimated constraint boundary.

If the estimator is
one-sided, one application of the error bound is saved and the gap
halves to $\epsilon/L_i$, but the shrinkage is still required in one of
the two cases. If the estimator only \emph{overestimates},
$\Lambda_i \le \widehat\Lambda_i \le \Lambda_i + \epsilon$, the unshrunk
rule $\widehat\Lambda_i(\widetilde W_i) \le \beta$ satisfies the true budget without reducing $\beta$,
since $\Lambda_i(\widetilde W_i) \le \widehat\Lambda_i(\widetilde W_i)
\le \beta$, and the binding case gives $\Lambda_i(\widetilde W_i) \ge
\beta - \epsilon$, hence $W_i^\star - \widetilde W_i \le \epsilon/L_i$.
If it only \emph{underestimates}, $\Lambda_i - \epsilon \le
\widehat\Lambda_i \le \Lambda_i$, the unshrunk rule can violate the true budget
(with $\Lambda(W) = W$ and $\widehat\Lambda(W) = \max(W - \epsilon, 0)$
it selects $W = \beta + \epsilon$), the shrinkage by $\epsilon$ must be
kept, and then $\Lambda_i(\widetilde W_i) \ge \widehat\Lambda_i(\widetilde
W_i) = \beta - \epsilon$ gives the same $\epsilon/L_i$ gap. We have not
measured the sign of the estimator error in training.
\end{remark}

\section{Supporting numerical diagnostics}
\label{app:diagnostics}

\subsection{Coverage tables and measurement convention}

The full-matrix estimator in Section~\ref{sec:estimators} measures
maximal endpoint stretching, whereas a random tangent vector measures
one direction. At $\tau=T_\lambda/2$ on L63, the directional proxy
has mean $1.09$, CV $1.87$, and $32\%$ non-positive values; the
full-matrix estimator gives $2.47$, $0.75$, and $0.2\%$.
Replacing directional stretch by maximal stretch also lowers CV from
$1.40$ to $0.53$ on R\"ossler, $1.41$ to $0.12$ on KS, and
$0.42$ to $0.11$ on L96, with no non-positive values on those systems
at this horizon. Random orientation therefore contributes substantial
variation that should not be interpreted as local instability
heterogeneity. All FTLE-based results reported here use the full
tangent matrix. Table~\ref{tab:cor3} in
Section~\ref{sec:pilot:ftle} gives the coverage results and
Table~\ref{tab:clipsweep} the floor-sweep results, both for $4000$
samples per system.

\begin{table}[t]
\centering
\small
\caption{Sensitivity of the realized gap to the instability floor at
$\tau = T_\lambda/2$, computed from the saved FTLE tables. Columns give
the percentile of positive FTLE values used as the floor $L$; entries
are the realized gap $R$ with the lower bound $\CVeff^2$ in parentheses. The
last column gives the fraction of samples with non-positive FTLE.}
\label{tab:clipsweep}
\begin{tabular}{lcccccc}
\toprule
System & $L$ at 1\% & 5\% & 10\% & 25\% & 50\% & non-positive FTLE \\
\midrule
L63 & $9.16\ (0.567)$ & $7.83\ (0.560)$ & $6.82\ (0.543)$ & $5.01\ (0.452)$ & $3.31\ (0.250)$ & $0.2\%$ \\
R\"ossler & $2.84\ (0.282)$ & $2.82\ (0.280)$ & $2.78\ (0.275)$ & $2.68\ (0.261)$ & $1.86\ (0.127)$ & $0.0\%$ \\
L96 ($N{=}40$) & $0.56\ (0.011)$ & $0.55\ (0.011)$ & $0.55\ (0.010)$ & $0.53\ (0.008)$ & $0.49\ (0.005)$ & $0.0\%$ \\
KS ($L{=}22$) & $0.56\ (0.014)$ & $0.56\ (0.014)$ & $0.56\ (0.014)$ & $0.55\ (0.013)$ & $0.53\ (0.011)$ & $0.0\%$ \\
\bottomrule
\end{tabular}
\end{table}

\subsection{True-system endpoint stretching}
\label{sec:pilot:lemma1}

The true-system endpoint stretch
$\Lambda^{\rm end}(W)=\log\sigma_{\max}(D\varphi^W)$, the
log of the largest singular value of the tangent matrix of the true
flow after time $W$, differs from
the nonnegative, non-decreasing cumulative quantity in
Lemma~\ref{lem:sensitivity}. For a flow satisfying
Assumption~\ref{assump:bounds},
\[
\log\|\Phi_i(W,0)\| \le \log C_\Phi+\Lambda_i(W).
\]
The prefactor cannot generally be omitted, and monotone endpoint growth
does not imply equality. Endpoint stretch can be negative or decrease
with horizon; every sampled L63 trajectory has at least one decrease.
These true-system measurements therefore describe finite-time stretching
without testing the learned-model bound. We integrate $20$ joint state
and tangent-matrix trajectories per system to $2T_\lambda$ and compare
their means and standard deviations at six horizons with the reference
line $\lambda_1^{\rm ref}W=W/T_\lambda$ (Table~\ref{tab:lemma1}).

\begin{table}[t]
\centering
\small
\caption{Endpoint tangent stretch on the true vector field:
trajectory-mean $\Lambda^{\rm end}(W) \pm$ standard deviation across $20$
attractor trajectories per system, with the system's $\CV$ at $\tau =
T_\lambda/2$ in parentheses. The bottom row is the reference line
$\lambda_1^{\rm ref} W = W/T_\lambda$.}
\label{tab:lemma1}
\begin{tabular}{lcccccc}
\toprule
System ($\CV$) & $W/T_\lambda{=}0.1$ & $0.25$ & $0.5$ & $1.0$ & $1.5$ & $2.0$ \\
\midrule
L63 ($0.75$) & $0.37 \pm 0.37$ & $0.88 \pm 0.66$ & $1.38 \pm 0.85$ & $2.17 \pm 1.15$ & $2.56 \pm 0.89$ & $3.14 \pm 1.07$ \\
R\"ossler ($0.53$) & $0.56 \pm 0.75$ & $0.83 \pm 0.62$ & $1.25 \pm 0.62$ & $1.86 \pm 0.61$ & $2.19 \pm 0.72$ & $2.53 \pm 0.96$ \\
L96 ($N{=}40$) ($0.11$) & $0.43 \pm 0.03$ & $0.93 \pm 0.07$ & $1.63 \pm 0.17$ & $2.61 \pm 0.35$ & $3.31 \pm 0.54$ & $3.98 \pm 0.53$ \\
KS ($L{=}22$) ($0.12$) & $1.00 \pm 0.07$ & $1.76 \pm 0.10$ & $2.43 \pm 0.16$ & $3.30 \pm 0.63$ & $3.79 \pm 0.55$ & $4.24 \pm 0.93$ \\
\midrule
\emph{Reference line} $\lambda_1^{\rm ref} W$ & $0.10$ & $0.25$ & $0.50$ & $1.00$ & $1.50$ & $2.00$ \\
\bottomrule
\end{tabular}
\end{table}

\begin{figure}[t]
\centering
\includegraphics[width=0.95\textwidth]{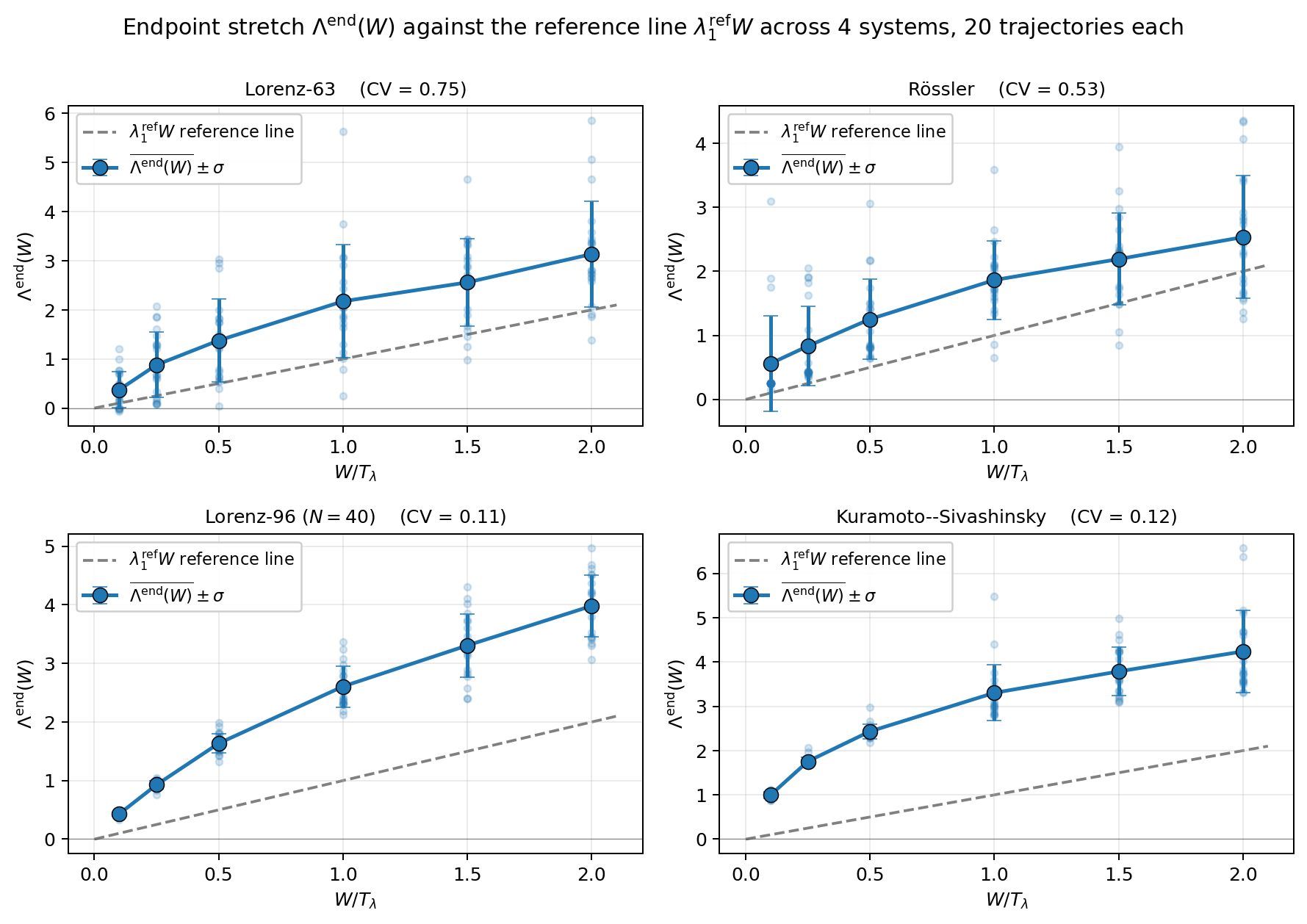}
\caption{Endpoint tangent stretch on the true vector field: mean
$\Lambda^{\rm end}(W)$ across $20$
attractor trajectories per system (points with
$\pm 1\sigma$ error bars) against the reference line $\lambda_1^{\rm ref} W$
(dashed). The finite-horizon slopes and offsets are descriptive;
this figure does not test the learned-model sensitivity bound. All
$20/20$ trajectories integrated successfully under the joint state and
tangent step.}
\label{fig:lemma1}
\end{figure}

Between $T_\lambda$ and $2T_\lambda$, mean stretch increases by
$0.97$, $0.67$, $1.37$, and $0.94$ on L63, R\"ossler, L96,
and KS, respectively, versus $1.0$ for the reference line. The means
remain above that line over the measured horizons
(Figure~\ref{fig:lemma1}); these finite-time increments and offsets
neither establish asymptotic slopes nor isolate non-normality as a cause.
At $W=T_\lambda$, the standard deviations are $1.15$, $0.61$,
$0.35$, and $0.63$, respectively. This disjoint sample agrees with
the CV ordering at its extremes, but not between R\"ossler and KS,
whose similar spreads contrast with CVs of $0.53$ and $0.12$.
With four systems, this is a descriptive comparison.

\subsection{Robustness to FTLE estimation error}
\label{sec:pilot:robust}

We test Proposition~\ref{prop:robust} on all four benchmark systems.
Starting from a true-system FTLE table at $\tau = T_\lambda/2$ ($n{=}2000$
attractor samples per system, floored at the $10$th percentile of
positive values as in Section~\ref{sec:pilot:coverage}, so that every
effective rate satisfies $\mu_i \ge L > 0$), we treat the table as the
true rate and set $\beta = \tau\bar\lambda_{\FTLE}$, which gives
$\beta = 1.43$, $1.36$, $1.62$ and $2.54$ on the four systems. Let
$W_{\max} = \beta/L$ be the largest horizon in the oracle schedule. We
perturb each rate by an error drawn uniformly from
$[-\epsilon/W_{\max}, \epsilon/W_{\max}]$ and truncate the result below
at $L$, so that the estimated profile $\widehat\Lambda_i(W) = \hat\mu_i
W$ satisfies $|\widehat\Lambda_i(W) - \Lambda_i(W)| \le \epsilon$ for
every admissible $W$, which is exactly the hypothesis of the
proposition, and build the budget-shrunken schedule $\widetilde
W_i(\beta;\epsilon)$. We measure (i) the fraction of segments that
remain feasible under the \emph{true} $\Lambda_i$
($\Lambda_i(\widetilde W_i) \le \beta$), and (ii) the realized maximum
coverage loss $C(W^\star) - C(\widetilde W)$ across $20$ independent
error realizations, against the bound $2\epsilon \sum_i 1/L_i$ with
$L_i = \mu_i$. We sweep $\epsilon \in \{0.02, 0.05, 0.1, 0.2, 0.4,
0.8\}$, and $1.6$ where it is below $\beta$, from about $1\%$ to over
$50\%$ of the budget. As a stress test we repeat the sweep with
Gaussian rate error of standard deviation $\epsilon/(2W_{\max})$,
followed by the same floor at $L$. Before flooring, the induced profile
error at $W_{\max}$ has standard deviation $\epsilon/2$; thus $\epsilon$
is a two-sigma level rather than a uniform bound. The untruncated
Gaussian perturbation exceeds this level in absolute value with
probability about $4.6\%$.

\begin{figure}[t]
\centering
\includegraphics[width=0.95\textwidth]{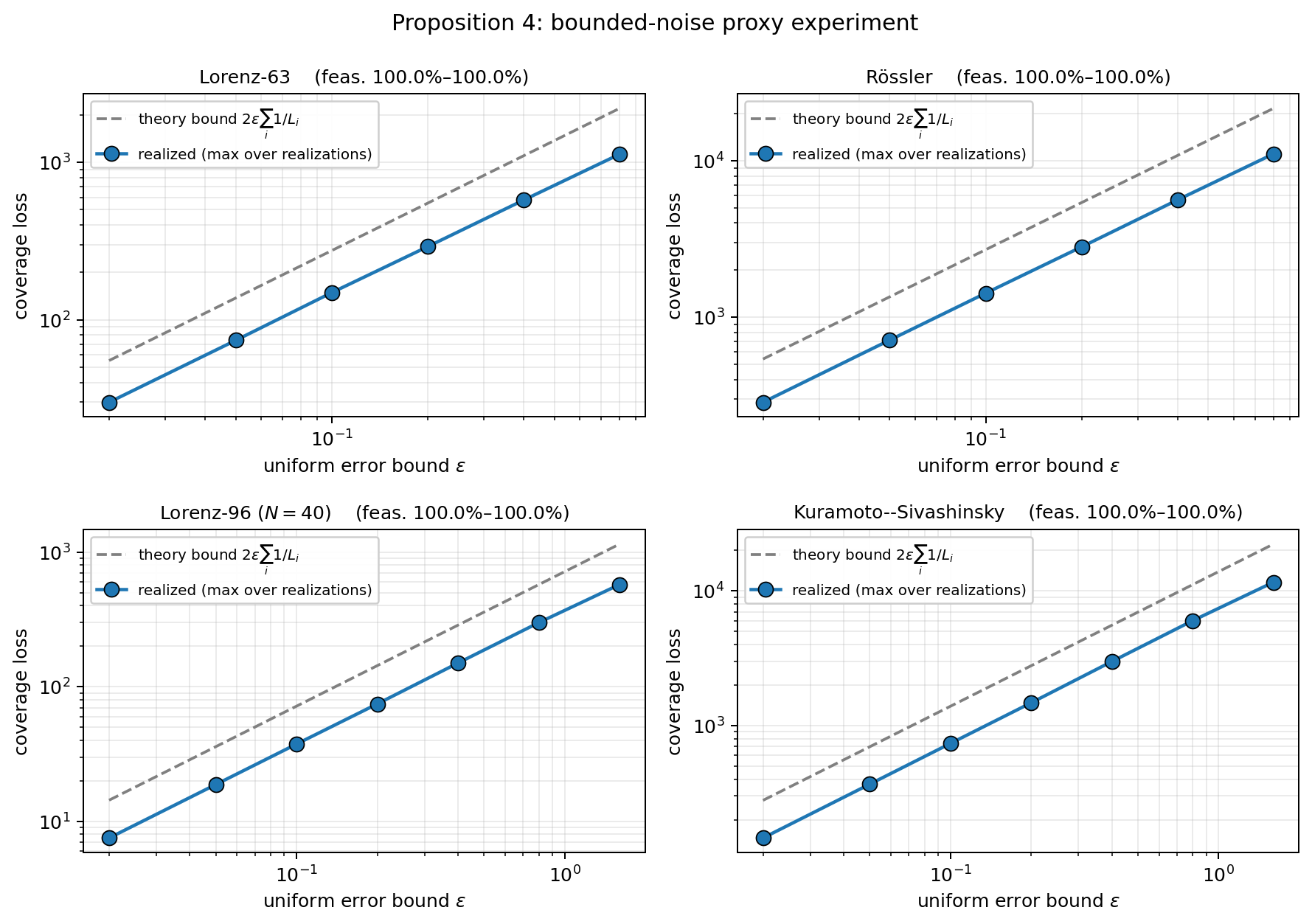}
\caption{Verification of Proposition~\ref{prop:robust} under bounded
error. Realized maximum coverage loss $C(W^\star) - C(\widetilde W)$
(solid markers) vs.\ the uniform error bound $\epsilon$, plotted
alongside the theoretical bound $2\epsilon \sum_i 1/L_i$ (dashed) on
log--log axes, one panel per system. Realized loss is $50$ to $54\%$ of
the bound at every error level on every system. Feasibility, annotated
in the panel titles, is $100\%$ in every cell.}
\label{fig:prop4}
\end{figure}

All $26$ bounded-error cells remain feasible, and realized coverage
loss is $50$--$54\%$ of the bound, tracking its linear scaling in
$\epsilon$ (Figure~\ref{fig:prop4}; Table~\ref{tab:prop4}
in Section~\ref{sec:pilot:ftle} summarizes the results). In the Gaussian stress test,
$0.05$--$1.6\%$ of segments violate the true budget at
$\epsilon\le0.4$, with more violations on the homogeneous systems,
whose horizons cluster near $W_{\max}$. No violations occur at
$\epsilon\ge0.8$, and the coverage-loss bound holds in every Gaussian
cell. These checks verify the proposition on linear true-system profiles;
they do not establish the accuracy of an estimator or a learned model.

\subsection{Per-segment gradient diagnostics}
\label{sec:study:lemma}

\begin{figure}[t]
\centering
\includegraphics[width=\textwidth]{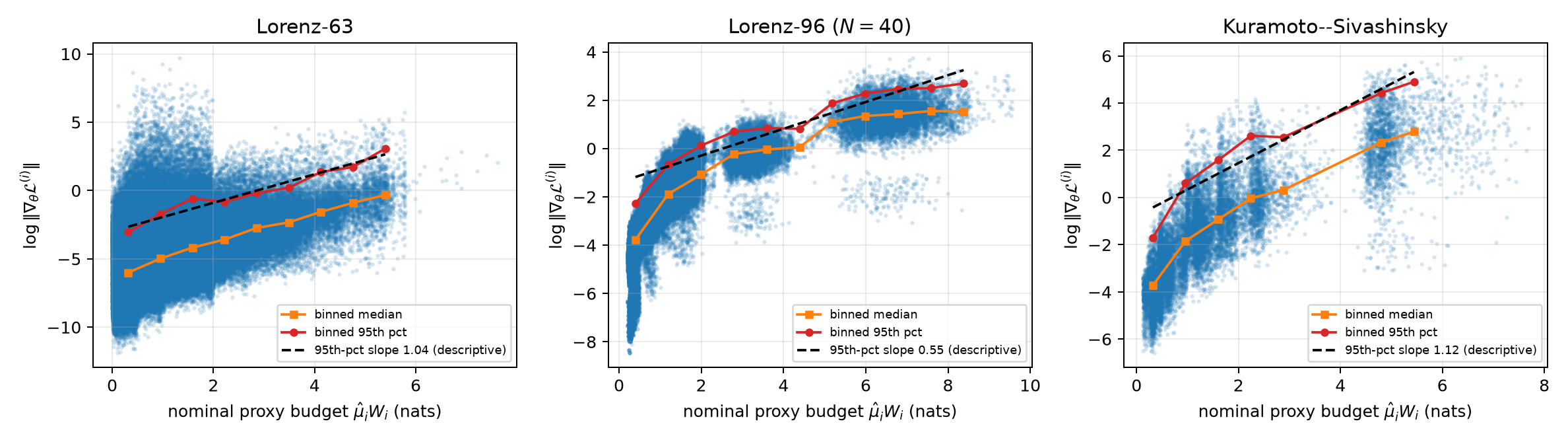}
\caption{Per-segment gradient norm against the nominal integrated
instability $\Lambda_i = \hat\mu_i W_i$ of the true-system profile, for
every segment logged during the runs with validation valid time above
$1\,T_\lambda$: $572{,}000$ segments on Lorenz-63 (the constant windows
of $1$ and $2\,T_\lambda$ excluded), $234{,}000$ on Lorenz-96, $15{,}000$
on KS. Orange: binned median; red: binned 95th percentile; dashed:
least-squares slope of the 95th-percentile curve. The slopes describe
how fast gradients grow with the nominal budget; they are not a test of
Lemma~\ref{lem:sensitivity}.}
\label{fig:lemma1seg}
\end{figure}

The $95$th-percentile gradient envelope rises at $1.04$ per nat on
L63, $0.55$ on L96, and $1.12$ on KS; medians rise more slowly, with
several nats of spread at a fixed nominal budget
(Figure~\ref{fig:lemma1seg}). These slopes describe gradients across
the included schedules. The horizontal axis is the true-system FTLE proxy
$\hat\mu_iW_i$, whereas Lemma~\ref{lem:sensitivity} uses learned-model
integrated instability and additional prefactors. An envelope slope
above or below one therefore neither confirms nor refutes that
pointwise bound.

\section{Full training-study tables}
\label{app:gridtables}

Tables~\ref{tab:grid_l63}--\ref{tab:grid_ks} list every schedule of the
training study, sorted by final gradient norm, with the realized
worst-case budget $\Lambda_{\max}$, final pre-clipping gradient norm,
valid time at both thresholds, first-failure time of $200\,T_\lambda$
rollouts (the ``First failure'' column, equal to $200$ when no failure occurs by the evaluation endpoint), and the error of the learned
Lyapunov exponent. Lyapunov errors are computed from complete
rollouts when no failure occurs, and from sufficiently long usable
initial parts when a rollout fails. Entries with error bars
report means $\pm$ population standard deviations across contributing
seeds. Prediction and failure-time summaries use all five seeds. Lyapunov-error summaries
use only seeds with a finite estimate from a rollout with a long enough usable part; in the
KS table, $n$ gives that metric's contributing seed count. Its mean and
population standard deviation are conditional on those seeds, and no
standard deviation is reported when $n=1$. Across KS, $47$ eligible
rollouts from $20$ runs contribute.

Compact schedule labels retain the budget and horizon information:
$S_w$ is a constant window $W=wT_\lambda$;
$C_\beta^c$, $A_\beta^c$, and $U_\beta^c$ denote the same-budget constant,
inverse-window-weighted adaptive, and unweighted adaptive rules,
respectively, at budget $\beta$ and cap $W_{\max}=cT_\lambda$.
Thus $c=0.5$ caps the horizon at $\tau$, whereas $c=2$ permits
extrapolation. $M_K$ uses $K$ chunks of a total $1T_\lambda$ window,
and $R$ is the global-Lyapunov ramp from $0.1$ to $1T_\lambda$.

\begin{table}[htbp]
\centering
\footnotesize
\caption{Lorenz-63, all $24$ schedules ($120$ runs).}
\label{tab:grid_l63}
\begin{tabular}{lcccccc}
\toprule
Schedule & $\Lambda_{\max}$ & $\|\nabla L\|$ & \shortstack{$t_{\rm valid}/T_\lambda$\\$(0.3)$} & \shortstack{$t_{\rm valid}/T_\lambda$\\$(0.5)$} & \shortstack{First failure\\$T/T_\lambda$} & $|\hat\lambda_1-\lambda_1|$ \\
\midrule
$C_{0.5}^{0.5}$ & $0.46$ & $9.2{\rm e}{-5}$ & $4.23 \pm 0.38$ & $4.68 \pm 0.42$ & $200$ & $0.011 \pm 0.005$ \\
$C_{0.5}^{2}$ & $0.46$ & $1.1{\rm e}{-4}$ & $4.24 \pm 0.14$ & $4.64 \pm 0.19$ & $200$ & $0.014 \pm 0.009$ \\
$C_{1}^{0.5}$ & $0.95$ & $3.3{\rm e}{-4}$ & $4.20 \pm 0.19$ & $4.66 \pm 0.20$ & $200$ & $0.018 \pm 0.007$ \\
$C_{1}^{2}$ & $0.95$ & $3.6{\rm e}{-4}$ & $4.41 \pm 0.09$ & $4.82 \pm 0.10$ & $200$ & $0.015 \pm 0.010$ \\
$S_{0.1}$ & $0.87$ & $4.6{\rm e}{-4}$ & $4.05 \pm 0.35$ & $4.50 \pm 0.32$ & $200$ & $0.014 \pm 0.007$ \\
$A_{0.5}^{2}$ & $0.49$ & $6.9{\rm e}{-4}$ & $4.07 \pm 0.41$ & $4.51 \pm 0.34$ & $200$ & $0.017 \pm 0.005$ \\
$U_{0.5}^{0.5}$ & $0.49$ & $7.0{\rm e}{-4}$ & $4.05 \pm 0.35$ & $4.47 \pm 0.24$ & $200$ & $0.014 \pm 0.013$ \\
$C_{2}^{0.5}$ & $1.96$ & $7.1{\rm e}{-4}$ & $4.16 \pm 0.37$ & $4.79 \pm 0.32$ & $200$ & $0.017 \pm 0.005$ \\
$A_{0.5}^{0.5}$ & $0.49$ & $8.5{\rm e}{-4}$ & $4.13 \pm 0.09$ & $4.77 \pm 0.18$ & $200$ & $0.019 \pm 0.015$ \\
$C_{2}^{2}$ & $1.93$ & $1.1{\rm e}{-3}$ & $4.36 \pm 0.30$ & $4.85 \pm 0.28$ & $200$ & $0.015 \pm 0.006$ \\
$M_{5}$ & $2.24$ & $1.3{\rm e}{-3}$ & $5.13 \pm 0.16$ & $5.59 \pm 0.20$ & $200$ & $0.021 \pm 0.003$ \\
$S_{0.25}$ & $2.15$ & $1.6{\rm e}{-3}$ & $4.49 \pm 0.17$ & $4.89 \pm 0.18$ & $200$ & $0.019 \pm 0.008$ \\
$A_{1}^{0.5}$ & $0.99$ & $2.2{\rm e}{-3}$ & $4.33 \pm 0.51$ & $4.80 \pm 0.47$ & $200$ & $0.018 \pm 0.002$ \\
$U_{1}^{0.5}$ & $0.98$ & $2.2{\rm e}{-3}$ & $4.15 \pm 0.32$ & $4.66 \pm 0.24$ & $200$ & $0.013 \pm 0.006$ \\
$U_{2}^{0.5}$ & $1.95$ & $4.1{\rm e}{-3}$ & $4.30 \pm 0.15$ & $4.73 \pm 0.23$ & $200$ & $0.018 \pm 0.005$ \\
$M_{3}$ & $3.48$ & $4.2{\rm e}{-3}$ & $5.08 \pm 0.24$ & $5.56 \pm 0.30$ & $200$ & $0.018 \pm 0.014$ \\
$A_{2}^{0.5}$ & $1.96$ & $5.0{\rm e}{-3}$ & $4.70 \pm 0.45$ & $5.17 \pm 0.34$ & $200$ & $0.016 \pm 0.009$ \\
$S_{0.5}$ & $4.69$ & $8.6{\rm e}{-3}$ & $4.72 \pm 0.22$ & $5.05 \pm 0.22$ & $200$ & $0.018 \pm 0.008$ \\
$M_{2}$ & $4.98$ & $4.6{\rm e}{-2}$ & $4.15 \pm 0.14$ & $4.64 \pm 0.19$ & $200$ & $0.018 \pm 0.010$ \\
$A_{1}^{2}$ & $0.97$ & $6.5{\rm e}{-2}$ & $2.67 \pm 0.16$ & $3.09 \pm 0.17$ & $200$ & $0.020 \pm 0.009$ \\
$R$ & $5.33$ & $9.3{\rm e}{-2}$ & $4.41 \pm 0.33$ & $4.94 \pm 0.37$ & $200$ & $0.023 \pm 0.005$ \\
$A_{2}^{2}$ & $1.96$ & $9.8{\rm e}{-1}$ & $1.96 \pm 0.17$ & $2.33 \pm 0.17$ & $200$ & $0.011 \pm 0.006$ \\
$S_{1}$ & $8.07$ & $2.5{\rm e}{+0}$ & $0.54 \pm 0.36$ & $0.99 \pm 0.58$ & $200$ & $1.239 \pm 1.195$ \\
$S_{2}$ & $17.73$ & $1.0{\rm e}{+2}$ & $0.35 \pm 0.16$ & $0.64 \pm 0.16$ & $200$ & $1.622 \pm 0.443$ \\
\bottomrule
\end{tabular}

\end{table}

\begin{table}[htbp]
\centering
\footnotesize
\caption{Lorenz-96 ($N{=}40$), all $24$ schedules ($120$ runs).}
\label{tab:grid_l96}
\begin{tabular}{lcccccc}
\toprule
Schedule & $\Lambda_{\max}$ & $\|\nabla L\|$ & \shortstack{$t_{\rm valid}/T_\lambda$\\$(0.3)$} & \shortstack{$t_{\rm valid}/T_\lambda$\\$(0.5)$} & \shortstack{First failure\\$T/T_\lambda$} & $|\hat\lambda_1-\lambda_1|$ \\
\midrule
$S_{0.1}$ & $0.39$ & $2.4{\rm e}{-3}$ & $1.39 \pm 0.02$ & $1.96 \pm 0.02$ & $200$ & $0.095 \pm 0.030$ \\
$C_{0.5}^{0.5}$ & $0.44$ & $2.4{\rm e}{-3}$ & $1.37 \pm 0.01$ & $1.94 \pm 0.02$ & $200$ & $0.093 \pm 0.041$ \\
$C_{0.5}^{2}$ & $0.42$ & $2.6{\rm e}{-3}$ & $1.39 \pm 0.02$ & $1.96 \pm 0.02$ & $200$ & $0.091 \pm 0.031$ \\
$A_{0.5}^{0.5}$ & $0.47$ & $3.5{\rm e}{-3}$ & $1.37 \pm 0.01$ & $1.94 \pm 0.02$ & $200$ & $0.097 \pm 0.047$ \\
$U_{0.5}^{0.5}$ & $0.47$ & $3.6{\rm e}{-3}$ & $1.37 \pm 0.01$ & $1.94 \pm 0.02$ & $200$ & $0.079 \pm 0.037$ \\
$A_{0.5}^{2}$ & $0.48$ & $4.9{\rm e}{-3}$ & $1.39 \pm 0.02$ & $1.96 \pm 0.02$ & $200$ & $0.085 \pm 0.047$ \\
$C_{1}^{2}$ & $0.92$ & $9.7{\rm e}{-3}$ & $1.38 \pm 0.02$ & $1.95 \pm 0.02$ & $200$ & $0.068 \pm 0.036$ \\
$C_{1}^{0.5}$ & $0.93$ & $1.0{\rm e}{-2}$ & $1.39 \pm 0.02$ & $1.95 \pm 0.01$ & $200$ & $0.078 \pm 0.044$ \\
$S_{0.25}$ & $1.10$ & $1.3{\rm e}{-2}$ & $1.37 \pm 0.02$ & $1.97 \pm 0.02$ & $200$ & $0.101 \pm 0.057$ \\
$A_{1}^{2}$ & $0.96$ & $1.5{\rm e}{-2}$ & $1.37 \pm 0.03$ & $1.95 \pm 0.04$ & $200$ & $0.109 \pm 0.048$ \\
$A_{1}^{0.5}$ & $0.95$ & $1.6{\rm e}{-2}$ & $1.38 \pm 0.02$ & $1.95 \pm 0.02$ & $200$ & $0.081 \pm 0.034$ \\
$U_{1}^{0.5}$ & $0.95$ & $1.6{\rm e}{-2}$ & $1.37 \pm 0.02$ & $1.97 \pm 0.02$ & $200$ & $0.093 \pm 0.031$ \\
$M_{5}$ & $0.87$ & $2.1{\rm e}{-2}$ & $1.37 \pm 0.03$ & $1.94 \pm 0.02$ & $200$ & $0.099 \pm 0.029$ \\
$R$ & $1.64$ & $3.1{\rm e}{-2}$ & $1.38 \pm 0.01$ & $1.96 \pm 0.02$ & $200$ & $0.124 \pm 0.034$ \\
$C_{2}^{2}$ & $1.89$ & $3.4{\rm e}{-2}$ & $1.37 \pm 0.02$ & $1.95 \pm 0.02$ & $200$ & $0.074 \pm 0.029$ \\
$S_{0.5}$ & $2.19$ & $3.8{\rm e}{-2}$ & $1.39 \pm 0.01$ & $1.95 \pm 0.02$ & $200$ & $0.081 \pm 0.017$ \\
$C_{2}^{0.5}$ & $1.89$ & $4.7{\rm e}{-2}$ & $1.37 \pm 0.02$ & $1.95 \pm 0.02$ & $200$ & $0.098 \pm 0.038$ \\
$M_{3}$ & $1.49$ & $5.6{\rm e}{-2}$ & $1.38 \pm 0.02$ & $1.96 \pm 0.02$ & $200$ & $0.103 \pm 0.038$ \\
$A_{2}^{2}$ & $1.91$ & $5.6{\rm e}{-2}$ & $1.38 \pm 0.02$ & $1.97 \pm 0.03$ & $200$ & $0.099 \pm 0.042$ \\
$U_{2}^{0.5}$ & $1.89$ & $5.7{\rm e}{-2}$ & $1.38 \pm 0.01$ & $1.96 \pm 0.04$ & $200$ & $0.122 \pm 0.033$ \\
$A_{2}^{0.5}$ & $1.89$ & $5.8{\rm e}{-2}$ & $1.38 \pm 0.01$ & $1.98 \pm 0.01$ & $200$ & $0.108 \pm 0.060$ \\
$M_{2}$ & $2.20$ & $1.1{\rm e}{-1}$ & $1.39 \pm 0.02$ & $1.97 \pm 0.01$ & $200$ & $0.119 \pm 0.059$ \\
$S_{1}$ & $4.46$ & $1.5{\rm e}{-1}$ & $1.38 \pm 0.01$ & $1.94 \pm 0.01$ & $200$ & $0.142 \pm 0.047$ \\
$S_{2}$ & $9.00$ & $9.0{\rm e}{-1}$ & $1.30 \pm 0.01$ & $1.87 \pm 0.02$ & $200$ & $0.222 \pm 0.061$ \\
\bottomrule
\end{tabular}

\end{table}

\begin{table}[htbp]
\centering
\footnotesize
\caption{Kuramoto--Sivashinsky, all $13$ schedules ($65$ runs). The
``First failure'' column gives the first-failure time, the mean time at which the $200\,T_\lambda$
rollouts first cross the amplitude/non-finite failure criterion;
$n$ in the final column counts contributing seeds for the
Lyapunov error computed from the usable parts of rollouts before failure.}
\label{tab:grid_ks}
\begin{tabular}{lcccccc}
\toprule
Schedule & $\Lambda_{\max}$ & $\|\nabla L\|$ & \shortstack{$t_{\rm valid}/T_\lambda$\\$(0.3)$} & \shortstack{$t_{\rm valid}/T_\lambda$\\$(0.5)$} & \shortstack{First failure\\$T/T_\lambda$} & $|\hat\lambda_1-\lambda_1|$ \\
\midrule
$C_{0.5}^{0.5}$ & $0.47$ & $2.5{\rm e}{-3}$ & $2.17 \pm 0.21$ & $2.65 \pm 0.24$ & $17 \pm 7$ & $0.021$ ($n=1$) \\
$S_{0.1}$ & $0.70$ & $7.3{\rm e}{-3}$ & $2.21 \pm 0.13$ & $2.73 \pm 0.13$ & $28 \pm 19$ & $0.011 \pm 0.004$ ($n=2$) \\
$A_{0.5}^{0.5}$ & $0.48$ & $9.4{\rm e}{-3}$ & $2.26 \pm 0.12$ & $2.70 \pm 0.12$ & $18 \pm 8$ & $0.016$ ($n=1$) \\
$C_{1}^{0.5}$ & $0.96$ & $1.8{\rm e}{-2}$ & $2.17 \pm 0.17$ & $2.67 \pm 0.26$ & $18 \pm 9$ & $0.011$ ($n=1$) \\
$S_{0.25}$ & $1.75$ & $3.5{\rm e}{-2}$ & $2.21 \pm 0.24$ & $2.77 \pm 0.34$ & $25 \pm 16$ & $0.013 \pm 0.001$ ($n=2$) \\
$C_{2}^{0.5}$ & $1.96$ & $4.5{\rm e}{-2}$ & $2.29 \pm 0.13$ & $2.79 \pm 0.16$ & $17 \pm 7$ & $0.012$ ($n=1$) \\
$A_{1}^{0.5}$ & $0.97$ & $4.7{\rm e}{-2}$ & $2.31 \pm 0.10$ & $2.83 \pm 0.07$ & $19 \pm 9$ & $0.014$ ($n=1$) \\
$M_{5}$ & $1.48$ & $7.5{\rm e}{-2}$ & $2.36 \pm 0.10$ & $2.88 \pm 0.12$ & $17 \pm 5$ & $0.013 \pm 0.004$ ($n=2$) \\
$A_{2}^{0.5}$ & $1.94$ & $1.3{\rm e}{-1}$ & $2.25 \pm 0.18$ & $2.73 \pm 0.17$ & $19 \pm 10$ & $0.010 \pm 0.008$ ($n=2$) \\
$M_{3}$ & $2.47$ & $1.4{\rm e}{-1}$ & $1.80 \pm 0.86$ & $2.25 \pm 1.03$ & $13 \pm 5$ & $0.010$ ($n=1$) \\
$S_{0.5}$ & $3.54$ & $2.4{\rm e}{-1}$ & $2.30 \pm 0.13$ & $2.80 \pm 0.18$ & $19 \pm 8$ & $0.016 \pm 0.005$ ($n=3$) \\
$R$ & $3.44$ & $3.1{\rm e}{-1}$ & $2.31 \pm 0.13$ & $2.89 \pm 0.16$ & $21 \pm 12$ & $0.006 \pm 0.006$ ($n=2$) \\
$S_{1}$ & $7.21$ & $2.9{\rm e}{+0}$ & $2.16 \pm 0.08$ & $2.59 \pm 0.09$ & $19 \pm 14$ & $0.011$ ($n=1$) \\
\bottomrule
\end{tabular}

\end{table}

\clearpage

\section{Model-based proxy and bootstrapping alternatives}
\label{app:bootstrap}

When the true dynamics are unknown, local growth could instead be
estimated from the learned vector field,
\begin{equation}
\hat\mu_i(t_j) = \frac{1}{\Delta t}\,
\log \sigma_{\max}\!\left( I + \Delta t\, \partial_x f_\theta(\hat x_i(t_j)) \right),
\label{eq:jac-proxy}
\end{equation}
where $\hat x_i(t_j)$ is the model state along the rollout,
$I$ is the identity, and $\sigma_{\max}$ is obtained from a few power iterations on Jacobian-vector
products, detached from the computational graph. Early in training
$f_\theta$ is uninformative, so the proxy needs an initial phase
that uses the true system's FTLE or a
switch based on validation error, as described below.
Proposition~\ref{prop:robust} says how to use it once its error is
bounded. This proxy is not evaluated here.

One alternative to that initial phase blends true-system and model estimates as
$\hat\mu_i=(1-\alpha_k)\hat\mu_i^{\mathrm{true}}+
\alpha_k\hat\mu_i^{\mathrm{model}}$, with $\alpha_k$ increasing from
$0$ to $1$. This requires access to the true system. When the
dynamics are unknown, one could initially use a schedule based on the
global Lyapunov exponent. Once the one-step validation error falls below
a chosen threshold, the schedule could use estimates from the learned
model. None of these
alternatives was tested in this study; every training run uses true-system FTLEs
throughout.

\section{Training details}
\label{app:algorithm}

For each system one long post-transient orbit is sliced into
disjoint windows separated by gaps of $5\,T_\lambda$: a
$200\,T_\lambda$ training window, a $20\,T_\lambda$ validation window,
a $200\,T_\lambda$ reference window for long-time statistics, and $50$
test windows of $15\,T_\lambda$ each. The data step is $\Delta t=0.01$
on L63 and L96 and $0.05$ on KS. The training, validation, reference, and test windows were
checked for identical rows when the data were generated. The true-system FTLE at
$\tau=T_\lambda/2$ is computed at every fifth training point (every
tenth on KS) and interpolated.

Lorenz-63 uses a 4-layer MLP of width $128$ with SiLU activations;
Lorenz-96 and KS use a periodic one-dimensional convolutional field
(5 layers, 64 channels, kernel 5) to model periodic spatial coupling.
A dense MLP failed to learn Lorenz-96 at the same budget in preliminary
experiments.

Training uses AdamW with initial learning rate $10^{-3}$, weight decay
$10^{-5}$, batches of $64$ segments ($96$ on KS), and a minimum window
$W_{\min}$ of five model steps ($5\Delta t$; $10\Delta t$ on KS). The
validation rollout loss over $T_\lambda/2$ is evaluated every $100$
updates ($250$ on KS), and a checkpoint is replaced when this loss
improves by more than $1\%$ relative to the saved value. PyTorch's
\texttt{ReduceLROnPlateau} scheduler uses the same relative threshold,
patience $5$ validation checks ($4$ on KS), reduction factor $0.5$, and
learning-rate floor $10^{-5}$. Training stops after $40$ non-improving
validation checks ($20$ on Lorenz-96, $12$ on KS), but not before
$20{,}000$ updates ($15{,}000$ on Lorenz-96, $8{,}000$ on KS), with a
ceiling of $60{,}000$ ($50{,}000$; $30{,}000$). Beyond the L63
schedules discussed in Section~\ref{sec:study:classes}, $15$ further
runs across nine schedules reach the ceiling in one to three seeds; the
two KS runs that do so use $W=1T_\lambda$.

The reported KS results use separate integration of each rollout.
In an earlier evaluation, all three rollouts from a run stopped when one
failed. Repeating the evaluation with separate integration changed
schedule-mean failure times by at most $2.14T_\lambda$. Short-time valid
times were unchanged, and every tested KS rollout still failed before
$200T_\lambda$. For L63 and L96, the rollouts were evaluated together,
and no long-rollout failure was recorded.

For state dimension $d$, the normalized root-mean-square error is
\[
\NRMSE(t)=\left[\frac{1}{d}\sum_{a=1}^d
\left(\frac{x_{\theta,a}(t)-x^\star_a(t)}{s_a}\right)^2\right]^{1/2},
\qquad
s_a=\operatorname{std}_{\rm train}(x_a)+10^{-8},
\]
where the standard deviation uses the training trajectory and the
population convention. Algorithm~\ref{alg:adaptive} summarizes one
training run of the adaptive rule.

\begin{algorithm}[h]
\caption{Adaptive instability-budget training with the smaller cap.}
\label{alg:adaptive}
\begin{algorithmic}[1]
\Require Training and validation trajectories; precomputed true-system
FTLEs at $\tau=T_\lambda/2$; floor $L$; model step $\Delta t_{\rm m}$;
$W_{\min}=5\Delta t_{\rm m}$, $W_{\max}=\tau$; update ceiling $K$;
batch size $B$; validation and stopping settings listed above
\State Initialize Neural ODE parameters $\theta$ and saved validation loss $v_{\rm best}\gets\infty$
\For{$k = 0,\ldots,K-1$}
    \State Compute $\beta_k$ via~\eqref{eq:betaschedule}
    \State Sample $B$ segment start points $\{(n_i, s_i)\}_{i=1}^B$
    \For{$i = 1$ to $B$}
        \State Read the interpolated true-system FTLE and set
        $\hat\mu_i\gets\max\{\widehat\lambda_{\FTLE,i}(\tau),L\}$
        \State $W_i\gets\min\{W_{\max},\max\{W_{\min},\beta_k/\hat\mu_i\}\}$
    \EndFor
    \State Group the $W_i$ into $q=4$ quantile groups; assign each group its minimum $W_i$
    \State $N_i\gets\max\{5,\lfloor W_i/\Delta t_{\rm m}\rfloor\}$,
    $W_i\gets N_i\Delta t_{\rm m}$, $\bar W\gets B^{-1}\sum_i W_i$
    \State Audit $\max_i\hat\mu_iW_i$ against $\beta_k$, including minimum-window violations
    \State Roll out groups with equal $N_i$; set $\mathcal{L}^{(i)}_{\mathrm{roll}}$ to each segment's mean normalized squared error
    \State $\mathcal{L}_{\mathrm{roll}} \gets \tfrac{1}{B}\sum_{i=1}^B (\bar W / W_i)\,\mathcal{L}^{(i)}_{\mathrm{roll}}$
    \State Update $\theta$ using AdamW
    \If{validation is due}
        \State Evaluate validation rollout loss $v$ over $T_\lambda/2$ and update the learning-rate scheduler
        \If{$v$ is finite and $v<0.99v_{\rm best}$}
            \State $(v_{\rm best},\theta_{\rm best})\gets(v,\theta)$; reset the stopping counter
        \Else
            \State Increment the stopping counter
        \EndIf
    \EndIf
    \If{the minimum update count and stopping patience are reached}
        \State \textbf{break}
    \EndIf
\EndFor
\State \Return $\theta_{\rm best}$
\end{algorithmic}
\end{algorithm}

\end{document}